\documentclass[10pt,twocolumn,letterpaper]{article}

\usepackage[pagenumbers]{cvpr} 
\usepackage{graphicx}  
\usepackage{booktabs,arydshln}
\definecolor{cvprblue}{rgb}{0.21,0.49,0.74}
\usepackage[pagebackref,breaklinks,colorlinks,allcolors=cvprblue]{hyperref}

\def\paperID{728} 
\def\confName{CVPR}
\def\confYear{2026}

\title{StoryTailor: A Zero-Shot Pipeline for Action-Rich Multi-Subject Visual Narratives}

\author{Jinghao Hu, Yuhe Zhang$^{\ast}$, Guohua Geng, Kang Li, Han Zhang\\
College of Computer Science, Northwest University\\
{\tt\small hujinghao@stumail.nwu.edu.cn, zhangyuhe0601@nwu.edu.cn}
}

\begin{document}

\maketitle

\begin{abstract}
Generating multi-frame, action-rich visual narratives without fine-tuning faces a threefold tension: action text faithfulness, subject identity fidelity, and cross frame background continuity. We propose StoryTailor, a zero-shot pipeline that runs on a single RTX 4090 (24 GB) and produces temporally coherent, identity-preserving image sequences from a long narrative prompt, per subject references, and grounding boxes. Three synergistic modules drive the system: Gaussian-Centered Attention (GCA) to dynamically focus on each subject core and ease grounding-box overlaps; Action-Boost Singular Value Reweighting (AB-SVR) to amplify action-related directions in the text embedding space; and Selective Forgetting Cache (SFC) that retains transferable background cues, forgets nonessential history, and selectively surfaces the retained cues to build cross scene semantic ties. Compared with baseline methods, the experiments show that CLIP-T improves by up to 10–15\%, with DreamSim lower than strong baselines, while CLIP-I stays in a visually acceptable, competitive range. With a matched resolution and steps on a 24 GB GPU, inference is faster than FluxKontext. Qualitatively, StoryTailor delivers expressive interactions and evolving yet stable scenes. 
\end{abstract}    
\section{Introduction}

\begin{figure*}[t]
  \centering
  \includegraphics[width=\textwidth]{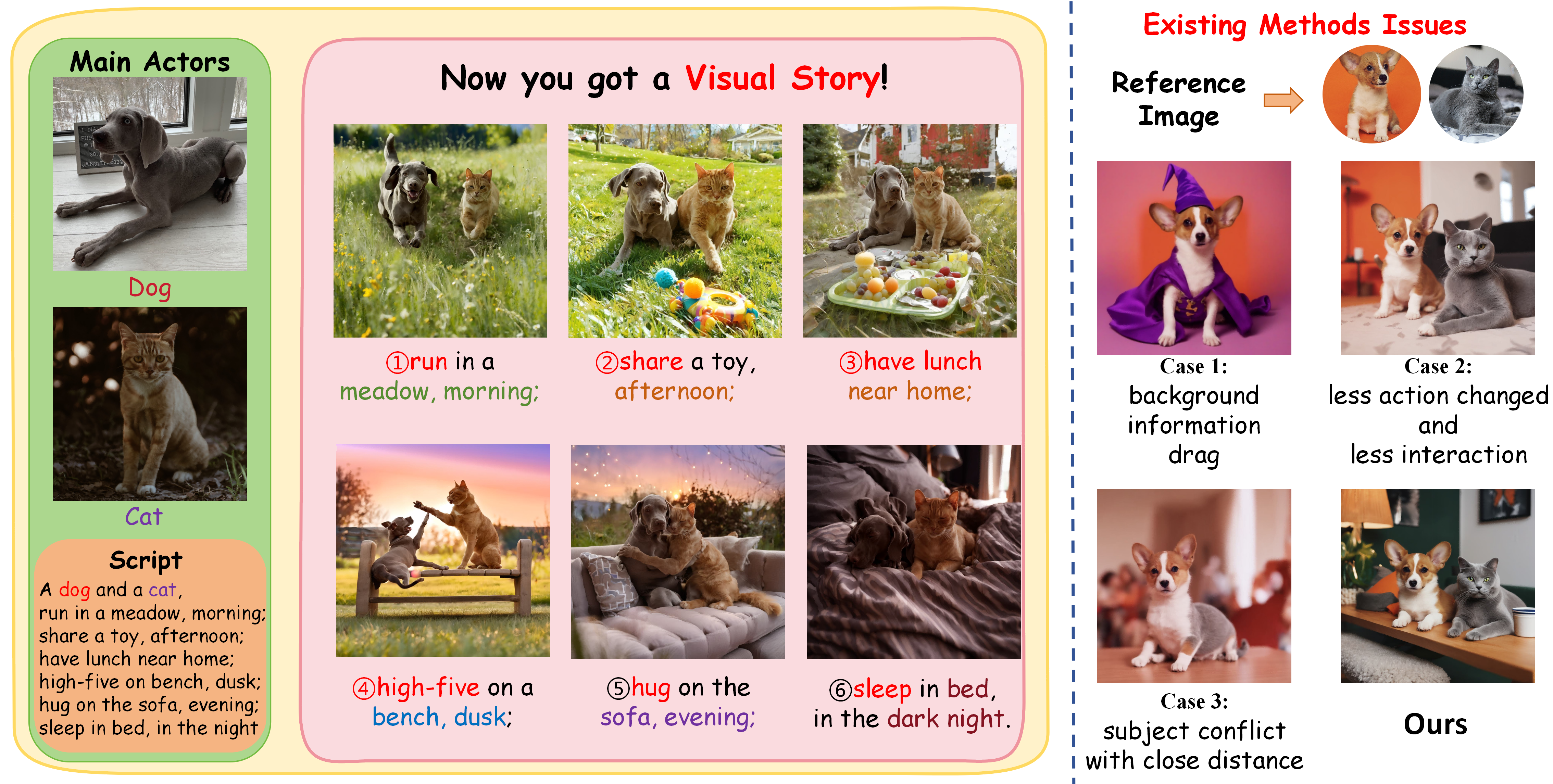}
  \caption{Zero-shot on a single 24-GB GPU, our pipeline produces action-rich multi-subject narratives with consistent identities and smooth scene transitions. Right: baselines show background drag, weak interaction, and close-range conflicts; our GCA stabilizes subjects, AB-SVR boosts actions, and SFC retains transferable background cues.}
  \label{fig:teaser}
\end{figure*}

Telling a story with pixels is easy; keeping who they are, what they do, and where they are consistent is not. Recent advances in multimodal vision language models\cite{radford2021learning} and diffusion models\cite{ho2020denoising,rombach2022high} have opened practical applications for personalized image and video generation across creative industries and interactive media. Identity consistent text-to-image and text-to-video (T2I/T2V)\cite{ye2023ip,wang2024ms,lin2024open,yuan2025identity,openai2024sora,chen2025multi} tasks are permeating applications in film preproduction and content creation, marketing and social media, games, XR, education and accessibility. Personalized narrative image synthesis extends identity consistent methods to a new setting: placing multiple characters in a shared scene and, under prompt guidance, generating multi-frame sequences that read as a story, while preserving identity fidelity, structuring clear interactions and actions, and maintaining cross frame scene continuity. 

Existing T2I personalization methods fall into two families: fine-tuning (e.g., DreamBooth \cite{ruiz2023dreambooth}, Textual Inversion \cite{gal2023textual}, LoRA \cite{hu2021lora}) and adapter methods (e.g., T2I-Adapter \cite{mou2023t2iadapterlearningadaptersdig}, IP-Adapter \cite{ye2023ip}, MS-Diffusion \cite{wang2024ms}). Fine-tuning typically requires multi-view, per-identity data and repeated optimization, which is costly and brittle; adapters are lighter but still largely single-frame. Both families often couple subjects with reference backgrounds, causing leakage outside regions, as shown in Fig.\ref{fig:teaser}.

Beyond personalization, the community explores sequence-level generators—video diffusion and in-context strategies \cite{openai2024sora,labs2025flux,comanici2025gemini,wu2025qwen}. These can improve temporal consistency but frequently require task-specific training, multi-view footage, or cluster-scale pipelines, which hinders reproducibility and deployment. Moreover, when subjects overlap or come into close physical proximity, in-context pipelines often entangle identities and fail to realize natural interactions, such as hugging, nestling.

To address the above limitations and align consistency modeling with narrative image synthesis\cite{ruiz2023dreambooth,hu2021lora,gal2023textual,ye2023ip,wang2024ms,labs2025flux,wu2025qwen,yang2024seed,wu2025diffsensei},we introduce StoryTailor, a zero-shot, multi-subject, interaction-oriented visual narrative pipeline. The system operates purely at inference on a single RTX 4090 (24 GB) and requires a long narrative prompt, persubject reference images, and grounding boxes as input. StoryTailor first integrates Resampler\cite{wang2024ms} into Stable Diffusion XL\cite{podell2023sdxl} to enable multi-subject conditioning and per subject grounding. Next, it applies Gaussian-Centered Attention (GCA). This softens boundaries inside each box while preserving subject centers, alleviating identity confusion from box overlaps and source-background carryover. To improve action controllability and behavioral diversity, an Action-Boost Singular Value Reweighting (AB-SVR) weights text-embedding components tied to actions. Cross frame scene continuity is maintained by a Selective Forgetting Cache (SFC) that propagates background context without constraining subject dynamics. Together, these designs deliver identity fidelity, interaction and action richness, and background consistency under practical compute.


The contributions can be summarized as follows:
\begin{itemize}

\item 
\textbf{Training-free multi-subject visual narratives.} A zero-shot diffusion-based pipeline on a single RTX 4090 (24 GB) that turns a long-form prompt, per subject references, and grounding boxes into action- and interaction-rich image sequences.

\item 
\textbf{AB-SVR.} Action-Boost Singular-Value Reweighting on text embeddings amplifies action and interaction directions; denoises other text directions, improving action controllability.

\item 
\textbf{GCA + SFC unified attention.} Couple GCA, a two-stage attention-guided Gaussian mask, with SFC, which uses correlation top-k to maintain transferable background cues. Together they stabilize subject centers and soften in-box boundaries, reducing subject confusion caused by close proximity and preserving continuity without freezing motion.

\end{itemize}

\section{Related Work}

\begin{figure*}[t]
  \centering
  \includegraphics[width=\textwidth]{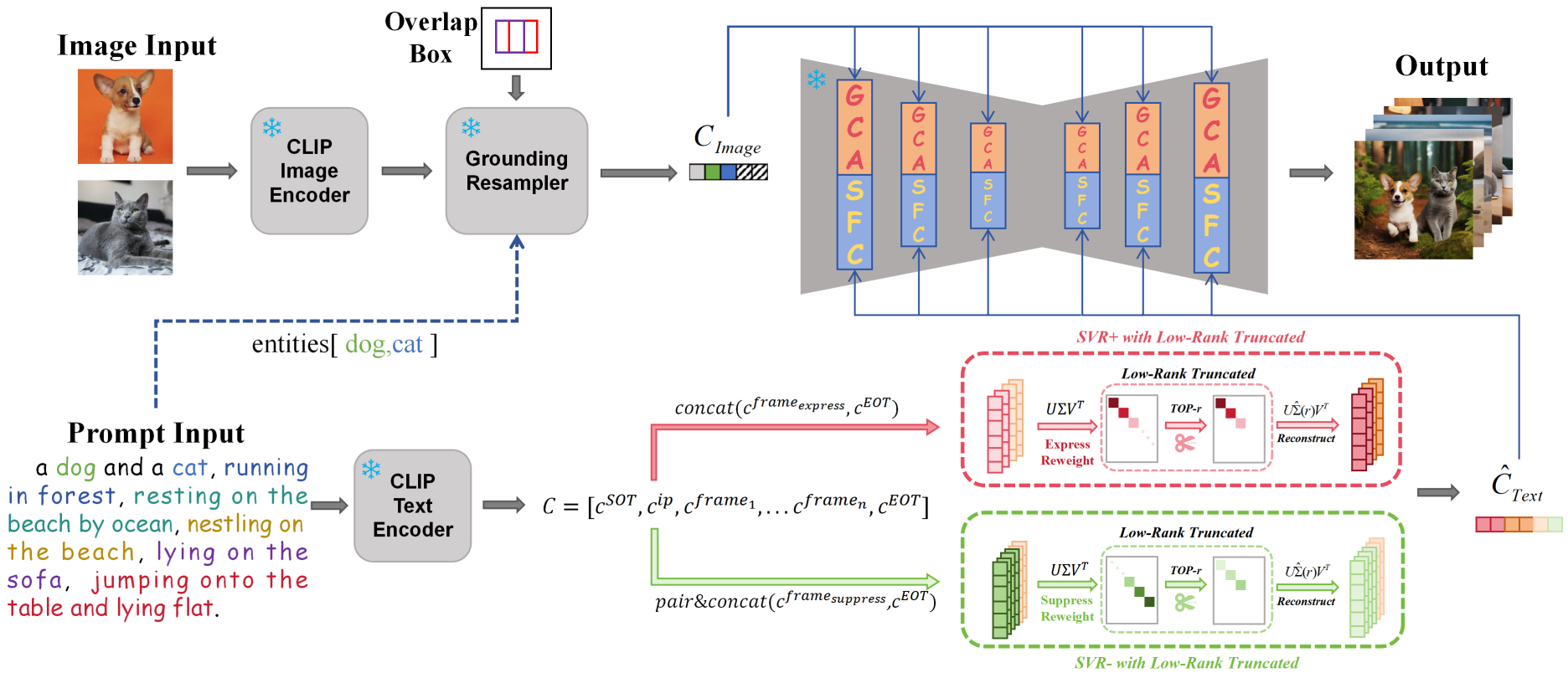}
  \caption{Overview of our pipeline: multi-subject conditioning and grounding, Gaussian-Centered Attention with Selective Forgetting Cache for layout recall, and AB-SVR on text embeddings to stabilize verbs and relations; frames are rendered sequentially into a coherent story.}
  \label{fig:pipeline}
\end{figure*}

\subsection{Consistency Methods}
Consistency methods were initially implemented through fine-tuning approaches, including techniques such as Textual Inversion\cite{gal2023textual}, LoRA\cite{hu2021lora}, and DreamBooth\cite{ruiz2023dreambooth}. Textual Inversion learns a pseudo-word embedding from a few images to bind an identity\cite{gal2023textual}; LoRA adds low-rank adapters that adjust attention or MLP weights with small parameter budgets\cite{hu2021lora}; DreamBooth performs full-parameter adaptation with class-prior regularization to retain identity\cite{ruiz2023dreambooth}. These early subject-consistency methods rely on per subject data and often imprint reference backgrounds and styles into outputs, causing scene-wide leakage.

Subsequent work shifted to adapter-based methods\cite{mou2023t2iadapterlearningadaptersdig,ye2023ip,wang2024ms}, which keep the backbone fixed and inject external guidance. IP-Adapter injects image features into cross-attention for zero-shot identity binding\cite{ye2023ip}, and MS-Diffusion augments multi-subject control with per subject references and grounding boxes\cite{wang2024ms}. Nonetheless, cross-attention in these methods tends to attend to reference backgrounds, overlapping boxes cause identity confusion, and action semantics remain weakly expressed and controlled.

Recent in-context pipelines\cite{wu2025qwen,labs2025flux,parihar2025kontinuous} (e.g., Flux-Kontext) treat text and images as a unified token stream, cast generation as context-conditioned prediction, and leverage large-scale data for general editing\cite{labs2025flux,parihar2025kontinuous}; however, they typically require GPU clusters and complex orchestration, and for consistency tasks, they struggle to realize multi-subject interactions. These gaps motivate our approach: explicitly localize attention on subjects and enhance the representation of actions in result images.

\subsection{Storytelling Methods}
The early storytelling pipelines\cite{ruiz2023dreambooth,hu2021lora,avrahami2024chosen} fixed the subject by fine-tuning (e.g. LoRA) and then generated frames one by one. This strategy offered quick personalization, but left weak linkage between frames and insufficient consistency protection, often manifesting as identity deformation, background leakage, and content drift. With the rise of GPU clusters and multimodal LLMs (MLLMs)\cite{radford2021learning,lin2023video}, methods such as SeedStory\cite{yang2024seed} and DiffSensei\cite{wu2025diffsensei,lin2023video} emerged: An MLLM parses and produces text–visual tokens, a decoder or diffusion model renders images, and optional LoRA-style tuning recovers subject identity\cite{yang2024seed,wu2025diffsensei}. These systems can produce full stories, yet styles skew toward animation, inference still demands about 33 GB VRAM with minute-level latency, and training typically relies on cluster-scale GPUs. Video synthesis models\cite{openai2024sora,deepmind2023imagen2} face similar hardware dependence and additionally suffer blur or ghosting, as seen in AnimateDiff\cite{guo2023animatediff}, under large viewpoint changes. Recent work\cite{liu2025one} (e.g., Liu et al.), inspired by FreeU\cite{si2024freeu,qin2025free}, introduces SVR at inference to enhance or suppress textual semantics, enabling story generation on consumer GPUs. However, identity preservation remains fragile, actions are limited, inter-frame relations weak, and customization is constrained\cite{liu2025one}. These limitations motivate our zero-shot design, which on a single 24 GB GPU combines explicit subject grounding, action-semantic reinforcement, and story-level coherence modeling to address these issues.
\section{Method}
Given a long narrative prompt, subject reference images, and grounding boxes, StoryTailor generates identity-preserving story sequences with two key modules. First,  Gaussian-Centered Attention(GCA) with a dynamic Gaussian distance-decayed mask loosens the restraints around the main body, softens box boundaries, and limits background carryover; a Selective Forgetting Cache forwards a controlled amount of background context to improve scene continuity without identity bleed. Second, Action-Boost Singular Value Reweighting reshapes the text embedding subspace by scaling high-energy action components, stabilizing verb semantics, and suppressing redundancy. Conditioned on these signals, the denoiser renders each frame in turn, yielding an interaction- and action-rich visual story. The StoryTailor pipeline is shown in Fig. \ref{fig:pipeline}.

\begin{figure*}[t] 
  \centering
  \includegraphics[width=\linewidth]{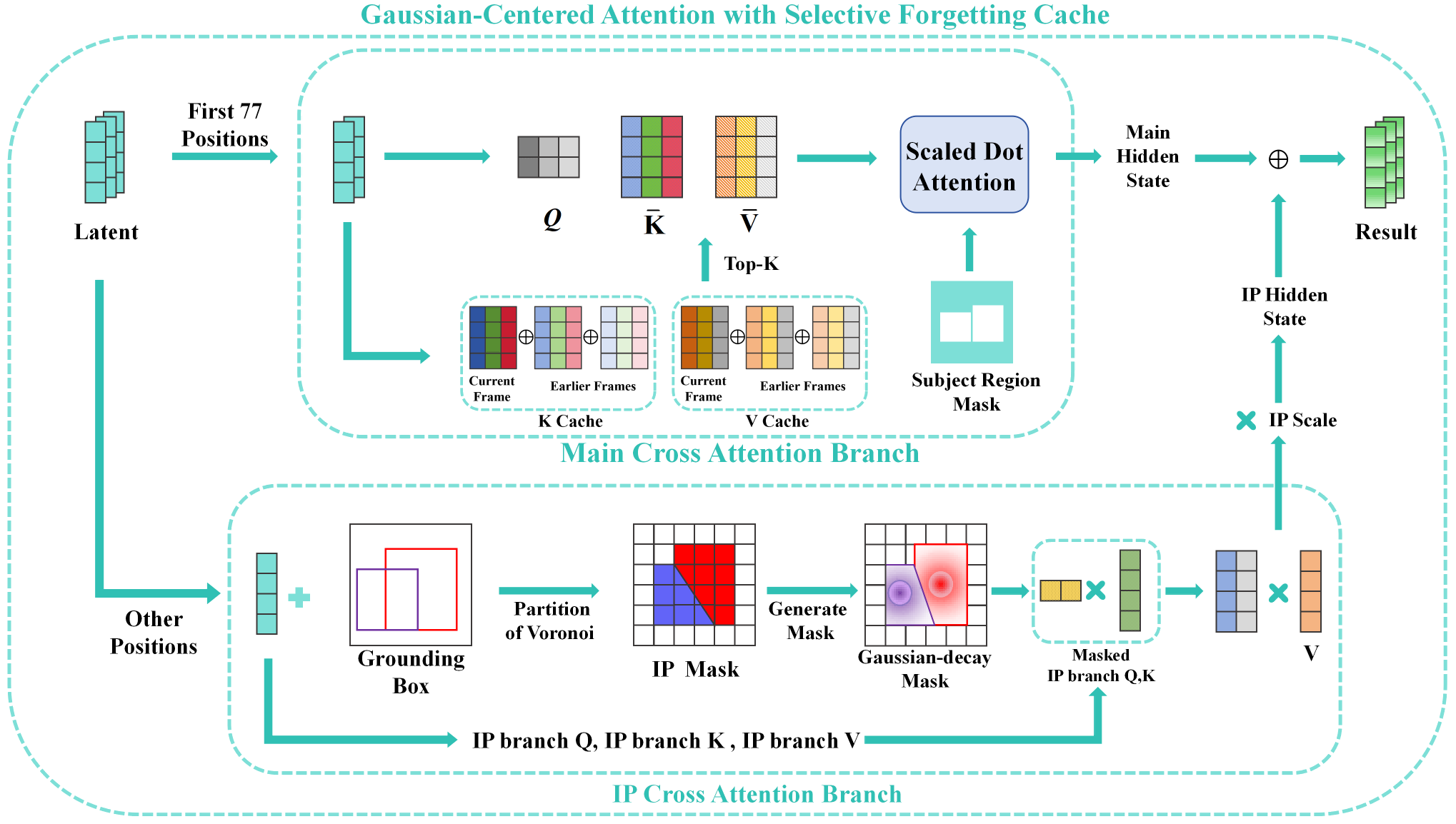}
  \caption{GCA with SFC: the first 77 tokens run the main cross-attention with a subject mask; the remaining tokens form an IP branch where overlaps yield a dynamic Gaussian-decay mask applied to Q,K. Cross frame K,V caches recall layout while limiting background carry-over, and the IP output is scaled and fused with the main state.}
  \label{fig:GCA}
\end{figure*}

\subsection{Preliminaries}
\subsubsection{Backbone}
We adopt SDXL\cite{podell2023sdxl} as the latent diffusion backbone\cite{rombach2022high} with MS-Diffusion\cite{wang2024ms} for multi-subject grounding. Text prompts are encoded into $c_t$ via SDXL encoder. Subject references are tokenized and resampled with boxes\cite{wang2024ms} into $c_i$. A VAE\cite{pinheiro2021variational} maps images to latents $z$, denoised by U-Net cross-attention, minimizing:
\begin{equation}
    L = \mathbb{E}{z,c,\epsilon,t} \left[ |\epsilon - \epsilon\theta (z_t \mid c_t, c_i, t)|^2_2 \right],
\end{equation}
where $z_t$ is noisy latent and $\epsilon \sim \mathcal{N}(0, I)$.

\subsubsection{Singular Value Reweighting(SVR)}
SVR\cite{liu2025one,si2024freeu,qin2025free,li2024get} modifies text embeddings $c_t$, segmented into tokens $C = [c_{\text{SOT}}, c_{P_0}, \dots, c_{P_N}, c_{\text{EOT}}]$. For target $P_j$, the express set (with $c_{P_j}$) is amplified, while suppress sets (others) are attenuated in embedding space. Updated $C'$ yields revised $\tilde{c}_t$ for cross-attention, boosting target expression and minimizing interference\cite{liu2025one}.

\subsection{Gaussian-Centered Attention}
Box-guided methods\cite{wang2024ms,wu2025diffsensei} often impose hard spatial boundaries; even with simple soft masks, attention still hugs the box edges, which constrains articulated motion and drags in irrelevant background. In practice, pushing for richer actions usually means cranking up text guidance, which hurts identity consistency through cross-attention drift\cite{wang2024ms,wu2025diffsensei}.

We therefore introduce Gaussian-Centered Attention (GCA): a center-initialized, two-stage attention-guided dynamic Gaussian-decay mask for cross attention. The GCA pipeline begins with the residual output of the previous stage's attention and proceeds as follows. We split the encoder states into text and IP tokens, running two cross-attention branches. For the text branch, with queries(merged head) $Q\in\mathbb{R}^{P\times d}$, keys and values $K_{\text{t}},V_{\text{t}}\in\mathbb{R}^{T_{\text{t}}\times d}$, we use scaled dot-product cross attention:
\begin{equation}
\alpha^{\text{text}}=\mathrm{softmax}\!\left(\frac{QK_t^\top}{\sqrt d}\right),\quad
H_{\text{text}}=\alpha^{\text{text}}V_t .
\end{equation}
For the IP branch, we apply a Gaussian-centered mask with a dynamic decay radius adjusted based on subject-specific factors, such as action intensity or overlap risk, to localize attention around subject centers while softening boundaries. To compute the mask, we first determine centers for each grounding box. For a box with coordinates $(x_1, y_1, x_2, y_2)$ on a patch grid, we use a Voronoi strategy to split and obtain sub-regions and compute their centroids $\mu_i^{\star}$. Decay radii $r_1$ and $r_2$ are initialized as fractions of the box's shorter side (base $r_1=0.35$, $r_2=0.70$). We then compute the average attention in the subject region from the text branch to obtain influence strength $p_i$, clipped to $[0,1]$ and mapped to a radius interval $[\sigma_{\min}, \sigma_{\max}]$ for the inner radius $s_{\text{in}}$. The outer radius is obtained via scaling factor $\rho > 1$:
\begin{equation}
s_{i}^{\text{in}} = \sigma_{\min} + (\sigma_{\max} - \sigma_{\min}) \, \operatorname{clip}(p_i, 0, 1) , s_{i}^{\text{out}} = \rho s_{i}^{\text{in}}.
\end{equation}
On the key grid $u=(x,y)$, we define anisotropic Gaussians for the inner and outer circles:
\begin{equation}
G^{\mathrm{in}}_i(u)=\exp\!\Big(-\tfrac{1}{2}\Big[\tfrac{(x-\mu_i^{\star x})^2}{s^2_{i,x,\mathrm{in}}}+\tfrac{(y-\mu_i^{\star y})^2}{s^2_{i,y,\mathrm{in}}}\Big]\Big),
\end{equation}
\begin{equation}
G^{\mathrm{out}}_i(u)=\exp\!\Big(-\tfrac{1}{2}\Big[\tfrac{(x-\mu_i^{\star x})^2}{s^2_{i,x,\mathrm{out}}}+\tfrac{(y-\mu_i^{\star y})^2}{s^2_{i,y,\mathrm{out}}}\Big]\Big),
\end{equation}
where $s_{i,x,\mathrm{in}}$ and $s_{i,y,\mathrm{in}}$ denote the anisotropic x- and y-axis scales of the inner radius $s_i^{\mathrm{in}}$; the outer ring is defined analogously. These are linearly fused and normalized to yield subject mask $M_i$, where inner slower decay protects identity core, and outer larger radius with faster decay decouples subjects from backgrounds, mitigating source background carryover. To avoid biasing irrelevant columns, $M_i$ is concatenated with dummy columns and broadcast across heads to form logit bias $B_{\text{ip}}$. Encoding IP hidden states yields $K_{\text{ip}}, V_{\text{ip}} \in \mathbb{R}^{(D + N_T) \times d}$, where $D$, $N_T$ as dummy token, text token. The IP branch attention is:
\begin{equation}
\alpha^{\text{ip}} = \operatorname{softmax}\!\left(
\frac{Q K_{\text{ip}}^{\top}}{\sqrt{d}} + B_{\text{ip}}
\right),
\qquad
H^{\text{ip}} = \alpha^{\text{ip}} V_{\text{ip}} .
\end{equation}
Finally, the IP-branch output is scaled by a preset subject factor and added back to the main branch to produce the final cross-attention update.

\begin{figure*}[t]
  \centering
  \includegraphics[width=\textwidth]{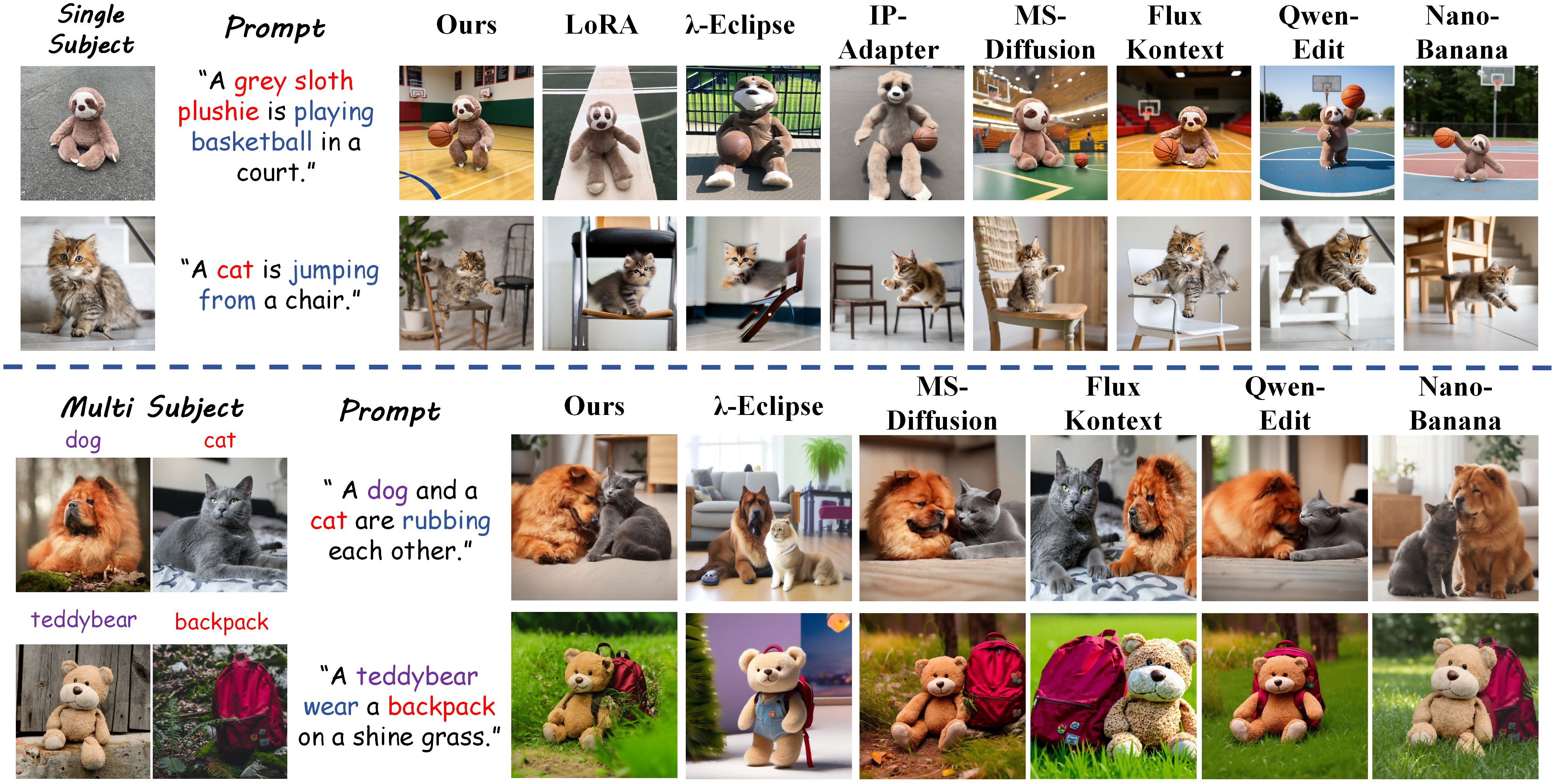}
  \caption{Single frame image consistency on single- and multi-subject tasks. Top: single-subject action sequences. Bottom: multi-subject interaction and attribute binding—dog and cat rubbing each other, teddy bear with a backpack. Only Ours and Qwen-Edit attach the backpack.}
  \label{fig:compare1}
\end{figure*}

\subsection{Action-Boost SVR}
Spatial unbinding alone is insufficient for stable, rich action semantics, and plain SVR only suppresses but does not zero the semantics of other frames, so residual action noise still interferes. We therefore operate in the text embedding space: denote the tokens of the current frame as $X_{exp}$ and the remaining frames as $X_{sup}$, and compute a thin SVD on the expressed block:
\begin{equation}
X_{\text{exp}} = U\,\Sigma\,V^\top,\qquad \Sigma=\mathrm{diag}(\sigma_1\ge\cdots\ge\sigma_m),
\end{equation}
where $U,V$ are the two sets of basis vectors for the SVD, $\sigma_i>0$ are the singular values of $X_{exp}$. Then select a kept rank by cumulative spectral energy to preserve only the principal “trunk” that carries actions:
\begin{equation}
   E(r)=\frac{\sum_{i=1}^{r}\sigma_i^2}{\sum_{i=1}^{m}\sigma_i^2},\qquad
k=\min\{\,r:\,E(r)\ge\tau\,\}, 
\end{equation}
let \(U_k=[u_1,\ldots,u_k]\) and \(P_k=U_kU_k^\top\), we keep the trunk for the current frame to stabilize action semantics and to prevent cross frame action leakage we notch-project the suppress frames away from this trunk:
\begin{equation}
\widetilde{X}_{\text{exp}} = P_k\,X_{\text{exp}}, 
\widetilde{X}_{\text{sup}}^{(\mathrm{notch})}=(I-P_k)\,X_{\text{sup}} ,
\end{equation}
where $I\!\in\!\mathbb{R}^{D\times D}$ is the identity; Finally, within the kept subspace we apply direction-selective, gentle SVR: the current frame is slightly enhanced while the others are attenuated, then both blocks are reconstructed and written back for subsequent denoising.

We visualize singular-value energy decays of text embeddings across multiple prompts and derive threshold bands around identifiable elbows and decay segments (Table~\ref{tab:tau-k-emphasis}); for action enhancement we set $\tau=0.85$ . Visualization results are provided in the Supplementary.
\begin{table}[t]
  \centering
    \caption{Cumulative-energy bands ($\tau$), typical $k$, and recommended $\tau$ by emphasis.}
  \setlength{\tabcolsep}{3pt}
  \renewcommand{\arraystretch}{1.02}
  \begin{tabular*}{\columnwidth}{@{\extracolsep{\fill}}l l c c@{}}
    \toprule
    \textbf{Emphasis} & \textbf{$\tau$ range} & \textbf{$k$ range} & \textbf{Rec.\ $\tau$} \\
    \midrule
    Identity / core shape      & 55--65\% & $2$--$3$   & 0.60 \\
    Actions / interactions     & 75--85\% & $3$--$4$   & 0.80 \\
    Background continuity      & 88--92\% & $5$--$6$   & 0.90 \\
    Style tone, lighting       & 92--95\% & $5$--$8$   & 0.93 \\
    Fine details, textures     & 97--99\% & $9$--$14$  & 0.98 \\
    \bottomrule
  \end{tabular*}
  \vspace{-2pt}
  \label{tab:tau-k-emphasis}
\end{table}

\subsection{Selective Forgetting Cache}
While GCA and AB-SVR ensure per-frame identity and action fidelity, cross frame background continuity is essential for coherent narratives but risks constraining subject dynamics if naively propagated.

We therefore introduce Selective Forgetting Cache (SFC), a diluted KV cache with adaptive forgetting that retains transferable background cues while discarding nonessential history. SFC operates in two synergistic modes: KV accumulation for cross attention and context mixing after SDPA; both are capacity capped to prevent memory explosion on a single 24 GB GPU.

For KV accumulation, given query \(Q \in \mathbb{R}^{B \times H \times P \times d_h}\), keys, values \(K,V \in \mathbb{R}^{B \times H \times T \times d_h}\) in a cross-attention layer excluding IP sources, we fetch historical \(K_h, V_h\) from a per-layer cache keyed by timestep, position, and batch. If in “accumulate” mode, we form the concatenated keys and values \(K_{\text{cat}} = [K_h'; K]\) and \(V_{\text{cat}} = [V_h'; V]\) along the sequence dimension, where \(K_h', V_h'\) are optionally filtered via top-\(k_h\) selection. We optionally select top-$k_h = 128$ tokens per query using unbiased similarity scores to ensure fair selection of relevant historical tokens before any forgetting bias is applied. The top-$k_h$ indices are computed as:
\begin{equation}
    \mathcal{I}_{\text{top}} = \arg\max_{k=1}^{k_h} \left( \frac{Q K_h^T}{\sqrt{d_h}} \right).
\end{equation}
The corresponding historical keys and values are gathered via $\mathcal{I}_{\text{top}}$ to form $K_h', V_h'$, preserving cues.
After concatenation, the attention logits over the historical segment are biased by $\delta_h = -0.1$ to promote gradual forgetting:
\begin{equation}
    S_{\text{hist}} = \frac{Q (K_h')^T}{\sqrt{d_h}} + \delta_h \cdot \mathbf{1}_{P \times k_h},
\end{equation}
where $\mathbf{1}_{P \times k_h}$ is an all-ones matrix. The full logits are then $S_{\text{full}} = [S_{\text{hist}}; S_{\text{current}}]$. To cap at $L_{\max} = 512$ tokens, we use FIFO decay (retain latest) or reservoir sampling (uniform downsample). Updated $K_{\text{cat}}, V_{\text{cat}}$ are written back unless in unconditional branches.

For post-SDPA context mixing, we cache the attention output $C$ per layer, averaged across batch to $\bar{C}$. At low-resolution layers, we mix previous $\bar{C}_{\text{prev}}$ into current $C$ only at background positions identified by mask $M_b \in \mathbb{R}^{B \times P}$:
\begin{equation}
    \tilde{C} = C \odot (1 - \alpha M_b') + \bar{C}_{\text{prev}} \odot (\alpha M_b'),
\end{equation}
where $M_b'$ is a subsampled version of the ratio of $M_b$, $\alpha = 0.6$ controls blend strength, $\odot$ denotes the Hadamard element-wise product, and shape mismatch is handled via nearest-neighbor resize. The unmixed $C$ is cached for the next frame, ensuring no feedback amplification. 

By selectively forgetting through capacity caps, history bias, decay, and top-k selection, and by surfacing through context mixing, SFC builds semantic ties across scenes without freezing motion and integrates seamlessly with GCA and AB-SVR for end-to-end narratives.

\section{Experiment}

\begin{figure*}[t]
  \centering
  \includegraphics[width=\textwidth]{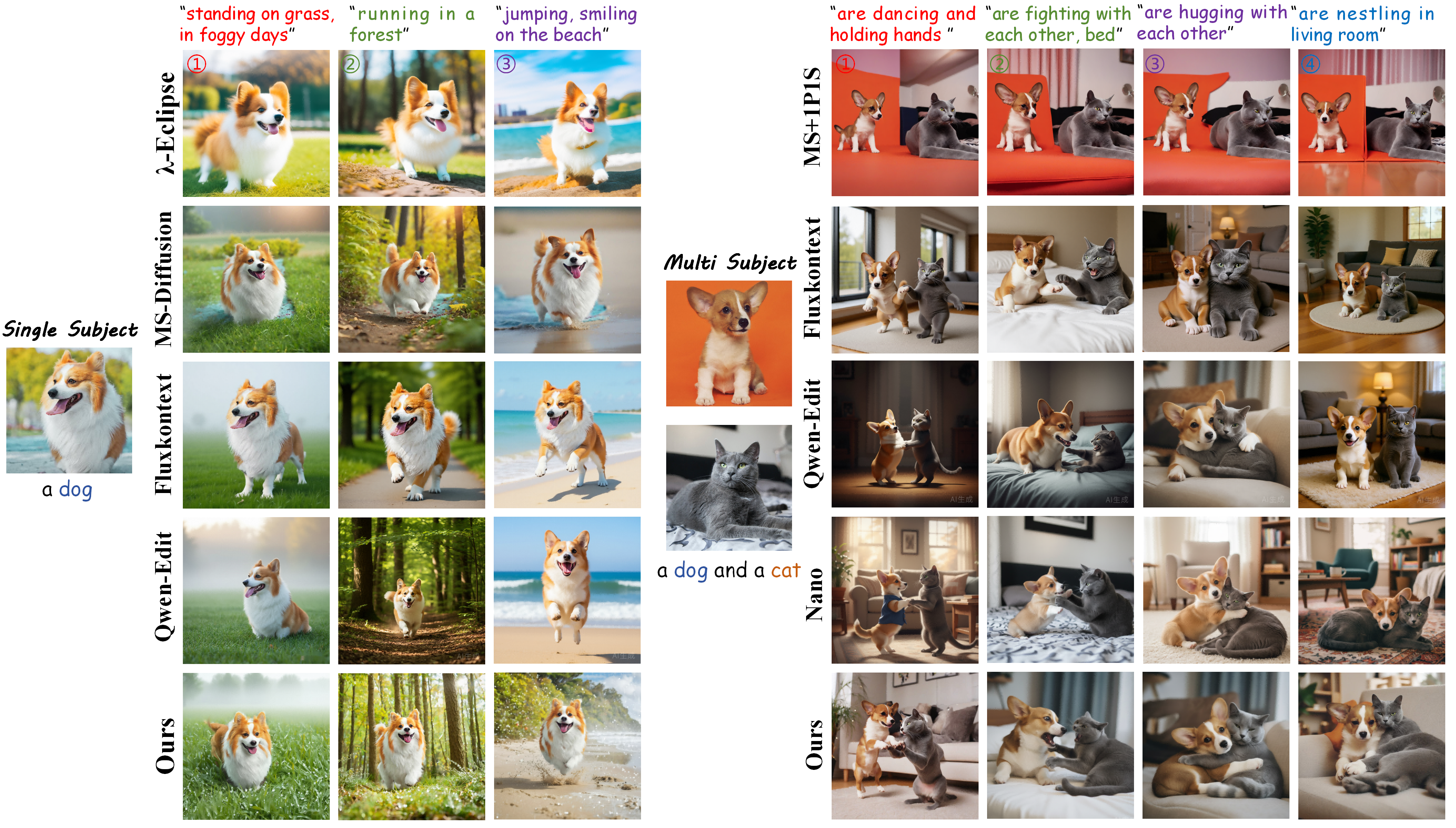}
  \caption{Prompt-following for actions and interactions in single- and multi-subject cases. With the same references and prompts, our method preserves identity and binds attributes to the correct subject while rendering actions and interactions—standing in fog, running in a forest, jumping on a beach; dancing and holding hands, play-fighting, hugging, nestling—with cleaner backgrounds than $\lambda$-Eclipse, MS-Diffusion, FluxKontext, Qwen-Edit, Nano-Banana, and MS+1P1S.}
  \label{fig:compare3}
\end{figure*}

\subsection{Experimental Setups}
\textbf{Benchmark and Tasks.} We evaluate on MSBench \cite{wang2024ms} and standardize all generation settings. We reuse the MSBench images but rewrite all prompts: for single-subject, we craft 100 long prompts per subject, each spanning 2–20 scenes; for multi-subject, we choose 40 interaction-capable subject combinations and create 100 long prompts for visual narratives. For every sequence we reconstruct the box locations for each scene according to the prompt. We report two regimes: single-frame consistency, and multi-frame visual narratives. We also run ablations on our three modules and compare related strategies. Unless otherwise noted, all results in this subsection use 30 inference steps with the same prompts, seeds, and evaluation protocol.

\newcommand{\best}[1]{\textbf{#1}}      
\newcommand{\second}[1]{\underline{#1}} 

\begin{table}[t]
  \centering
  \caption{Multi-subject image consistency on MSBench (micro-averaged means).
  \textbf{Best} in bold; \underline{second-best} underlined.}
  \label{tab:multi_subject_only}
  \setlength{\tabcolsep}{2.2pt}
  \renewcommand{\arraystretch}{1.03}

  \begin{tabular*}{\columnwidth}{@{\extracolsep{\fill}} lccc}
    \toprule
    \textbf{Method} & CLIP-I (↑) & M-DINO (↑) & CLIP-T (↑) \\
    \midrule
    \multicolumn{4}{l}{\textit{Adapters \& MLLM}}\\
    MS-Diffusion         & 0.692 & 0.108 & 0.340 \\
    SSR-Encoder          & 0.723 & 0.104 & 0.311 \\
    $\lambda$-ECLIPSE    & 0.719 & 0.097 & 0.304 \\
    \midrule
    \multicolumn{4}{l}{\textit{In-context}}\\
    FluxKontext          & \second{0.732} & 0.107 & 0.372 \\
    Qwen-Edit            & 0.714 & 0.108 & \second{0.396} \\
    Nano-Banana           & \best{0.749} & \best{0.114} & 0.389 \\
    \midrule
    \textbf{StoryTailor (Ours)} & 0.717 & \second{0.112} & \best{0.414} \\
    \bottomrule
  \end{tabular*}
\end{table}

\textbf{Metrics.}
CLIP-T ($\uparrow$): Text-image alignment via cosine similarity between prompt and image embeddings \cite{hessel2021clipscore}.

CLIP-I ($\uparrow$): Identity fidelity via cosine similarity between generated subject crops and reference crops \cite{hessel2021clipscore}.

DINOv2 ($\uparrow$): Appearance and identity consistency via cosine similarity of DINOv2 features from subject \cite{oquab2023dinov2}.

DreamSim ($\downarrow$): Perceptual coherence on full images and subject, lower scores indicating human-like similarity \cite{fu2023dreamsim}.

Inference Time ($\downarrow$): Mean end-to-end time per frame.

VRAM Peak ($\downarrow$):Inference maximum GPU memory.

\textbf{Baseline.}
We compare four families of methods: Textual Inversion\cite{gal2023textual}, LoRA\cite{hu2021lora} and DreamBooth\cite{ruiz2023dreambooth} as fine-tuning methods, MS-Diffusion\cite{wang2024ms} and IP-Adapter\cite{ye2023ip} as adapter methods, $\lambda$-Eclipse\cite{patel2024lambda} as an MLLM\cite{patel2024eclipse} method, and FluxKontext\cite{labs2025flux}, Qwen-Edit\cite{wu2025qwen}, Nano-Banana\cite{comanici2025gemini} as In-context methods. Notably, Qwen-Edit and Nano-Banana cannot be deployed on a RTX 4090 and are evaluated only via APIs; to avoid confounds from network response and system latency, we omit their latency metrics in both two tasks, while all accuracy metrics are still reported. We also include 1Prompt1Story\cite{liu2025one} as a consistency baseline by combining it with IP-Adapter and MS-Diffusion.

\begin{table*}[t]
  \centering
  \caption{Short-frame visual narratives including single and multi.
  Higher is better for CLIP-I and CLIP-T; lower is better for DreamSim.
  Efficiency is per frame on a single RTX 4090 (24 GB), 1024$\times$1024, 30 steps; API rows omit efficiency.}
  \label{tab:short_frame_clip_dreamsim_twocol}
  \small
  \setlength{\tabcolsep}{4pt} 
  \renewcommand{\arraystretch}{1.02}
  \resizebox{\textwidth}{!}{
    \begin{tabular}{ lccc c ccc c cc }
      \toprule
      & \multicolumn{3}{c}{\textbf{Single-subject}} &
      & \multicolumn{3}{c}{\textbf{Multi-subject}} &
      \multicolumn{2}{c}{\textbf{Efficiency}} \\
      \cmidrule(l{2pt}r{2pt}){2-4}
      \cmidrule(l{2pt}r{2pt}){6-8}
      \cmidrule(l{3pt}r{3pt}){9-10}
      \textbf{Method} & CLIP-I $\uparrow$ & CLIP-T $\uparrow$ & DreamSim $\downarrow$ & &
                         CLIP-I $\uparrow$ & CLIP-T $\uparrow$ & DreamSim $\downarrow$ &
                         Time (s) $\downarrow$ & VRAM (GB) $\downarrow$ \\
      \midrule
      IP-Adapter & 0.872 & 0.283 & 0.214 & & -- & -- & -- & \best{4.82} & \underline{19.39} \\
      IP-Adapter + 1Prompt1Story & \textbf{0.928} & 0.215 & 0.472 & & -- & -- & -- & 7.25 & 21.64 \\
      $\lambda$-Eclipse & 0.826 & 0.299 & 0.268 & & -- & -- & -- & 8.31 & \best{14.21} \\
      MS-Diffusion & 0.884 & 0.315 & 0.197 & & 0.687 & 0.318 & 0.323 & \underline{5.36} & 19.47 \\
      FluxKontext & 0.913 & 0.381 & 0.187 & & 0.732 & 0.373 & 0.212 & 32.31 & 22.99 \\
      Qwen-Edit & \underline{0.922} & \underline{0.392} & \textbf{0.169} & & 0.718 & 0.386 & 0.206 & -- & -- \\
      Nano-Banana & 0.911 & 0.389 & 0.183 & & \underline{0.737} & \underline{0.391} & \underline{0.195} & -- & -- \\
      MS-Diffusion + 1Prompt1Story & 0.917 & 0.232 & 0.296 & & \textbf{0.765} & 0.314 & 0.491 & 7.88 & 21.25 \\
      \textbf{Ours} & 0.869 & \textbf{0.431} & \underline{0.171} & & 0.671 & \textbf{0.406} & \textbf{0.189} & 7.02 & 20.01 \\
      \bottomrule
    \end{tabular}
  }
\end{table*}

\subsection{Single-frame Image Consistency}
\textbf{Qualitative.}
Fig.\ref{fig:compare1} presents visual results. Fine-tuning methods inherit training bias; and limited by multi-view subject data, motion expression and identity preservation weaken \cite{hu2021lora,gal2023textual,ruiz2023dreambooth}. IP-Adapter struggles to balance identity and textual guidance \cite{ye2023ip}; MS-Diffusion preserves identity yet shows conservative motion and occasional attribute spill \cite{wang2024ms}. In-context methods, without explicit guidance, often show background ambiguity or miss interactive actions \cite{labs2025flux,wu2025qwen,comanici2025gemini}. In contrast, StoryTailor, benefiting from GCA, preserves identity, renders crisp poses, keeps backgrounds clean, and binds attributes to the correct subject. 

\textbf{Quantitative.}
Table \ref{tab:multi_subject_only} shows clear CLIP-T gains. CLIP-I is slightly lower than baselines, attributable to richer actions and reduced background carry-over; this aligns with the qualitative observations. DINO remains strong, and single uses DINOv2(results in Supp.); multi uses M-DINO.


\subsection{Visual Narrative}
We present single- and multi-subject visual narrative results; some narrative examples are shown in Fig. \ref{fig:compare3}. Using a single prompt for narrative generation, our method preserves subject identity, produces diverse and controllable actions, and maintains scene continuity so that scenes and layouts remain similar across frames. 

\textbf{Qualitative.}
As shown in Fig. \ref{fig:compare3}, the figure presents a single-subject sequence on foggy grass, forest, and beach. GCA stabilizes the dog’s core while softly relaxing limb boundaries, preventing cross-box entanglement and keeping overlaps clean. AB-SVR strengthens verb tokens such as standing, running, and jumping so poses progress instead of collapsing into static portraits. SFC tracks background correlation and performs clean resets across outdoor scene changes to avoid carry-over; the right block shows a indoors sequence where it recalls transferable background cues to provide soft continuity. In the dog and cat interaction sequence covering dance and hand over, play fight, hug, and nestle, GCA limits tangling at contact points, AB-SVR makes the interactions reliably materialize, and SFC keeps the room coherent without freezing motion.

\begin{table}[t]
  \centering
  \caption{Ablation summary in Multi-subject Narratives task.
  Higher is better for CLIP-I/CLIP-T; lower is better for Time; Subject confuse(Y/N).}
  \label{tab:ablation_all_in_one}
  \setlength{\tabcolsep}{3pt}    
  \renewcommand\arraystretch{1.0}
    \resizebox{\linewidth}{!}{
   \small 
  \begin{tabular}{lcccc}
    \toprule
    \textbf{Variant} & \textbf{CLIP-I}$\uparrow$ & \textbf{CLIP-T}$\uparrow$ & \textbf{Time}$\downarrow$ & \shortstack{\textbf{Subject}\\\textbf{Confuse}} \\
    \midrule
    \multicolumn{5}{l}{\textit{Core modules}}\\[-2pt]
    Ours (GCA + SFC + SVR) & 0.671 & 0.406 & 7.026 & N \\
    w/o AB-SVR           & 0.636 & 0.364 & 6.715 & N \\
    w/o SFC                & 0.663 & 0.341 & 6.412 & N \\
    w/o GCA               & 0.684 & 0.319 & 5.894 & Y \\
    w/o GCA and SVR       & 0.675 & 0.304 & 5.431 & Y \\
    \bottomrule
  \end{tabular}
  }
\end{table}

\textbf{Quantitative.}
Table \ref{tab:short_frame_clip_dreamsim_twocol} indicates higher CLIP-T and lower DreamSim for both single- and multi-subject narratives, with a modest drop in CLIP-I. We attribute this to: AB-SVR amplifying verb and interaction directions, GCA stabilizing subject cores and softening in-box boundaries to reduce overlap entanglement, and SFC retaining transferable background cues while suppressing nonessential history, which mitigates background carry-over and attribute spill. Gains are more pronounced in multi-subject scenes with heavy box overlaps. Detailed numbers and subgroup analyzes are provided in the supplementary material.

\subsection{Ablation and Supplement}
As shown in Table \ref{tab:ablation_all_in_one}, the full model achieves the highest CLIP-T with minimal subject confusion in overlap regions, demonstrating stronger text–image alignment and cleaner cross-box boundaries. Removing AB-SVR markedly reduces CLIP-T, confirming its role in strengthening verb-related semantics; removing SFC lowers both CLIP-T and CLIP-I, consistent with weakened cross-frame context and identity stability; removing GCA slightly raises CLIP-I but decreases CLIP-T and introduces overlap confusion, showing that GCA limits background carry-over and boundary conflicts. Ablation visualizations are in Supplement.

We assess stability on 2–20-frame tasks under matched conditions. For GCA, we ablate anisotropy and partition schemes, including XOR Split, diffusion-guided Split, Static Gaussian, Single-stage etc.. Protocols and hyperparameter sweeps appear in the supplement. In addition, we run a 100-participant user study on interaction realism and identity stability; raters show a clear preference for the full model over ablated variants and baselines. As the number of frames grows, semantic noise accumulates in text embeddings: CLIP-T shows a mild decline beyond 16 frames, whereas CLIP-I and DreamSim remain stable. Overall, the selected GCA variant offers the best accuracy–efficiency trade-off, and the user study corroborates the visual gains.

\section{Conclusion}
We present StoryTailor, a pipeline for multi-subject, action-rich visual narratives that runs on a 24 GB GPU. Our method uses Gaussian-Centered Attention to reduce subject confusion and background drag, applies Action-Boost SVR in the text-embedding space to amplify verb semantics, and employs a Selective Forgetting Cache to inherit background context without suppressing subject dynamics. Empirically, we observe richer actions and clearer interactions with notably higher CLIP-T, while identity fidelity remains strong; the small drop in CLIP-I is acceptable. Overall, StoryTailor unifies spatial unbinding with cross frame context reuse into a practical, zero-shot solution for narrative synthesis, and opens avenues for finer micro-attribute control and adaptive KV inheritance under large scene transitions. 

{
    \small
    \bibliographystyle{ieeenat_fullname}
    \bibliography{main}
}

\clearpage
\setcounter{page}{1}
\maketitlesupplementary

\begin{figure*}[t] 
  \centering
  \includegraphics[width=\linewidth]{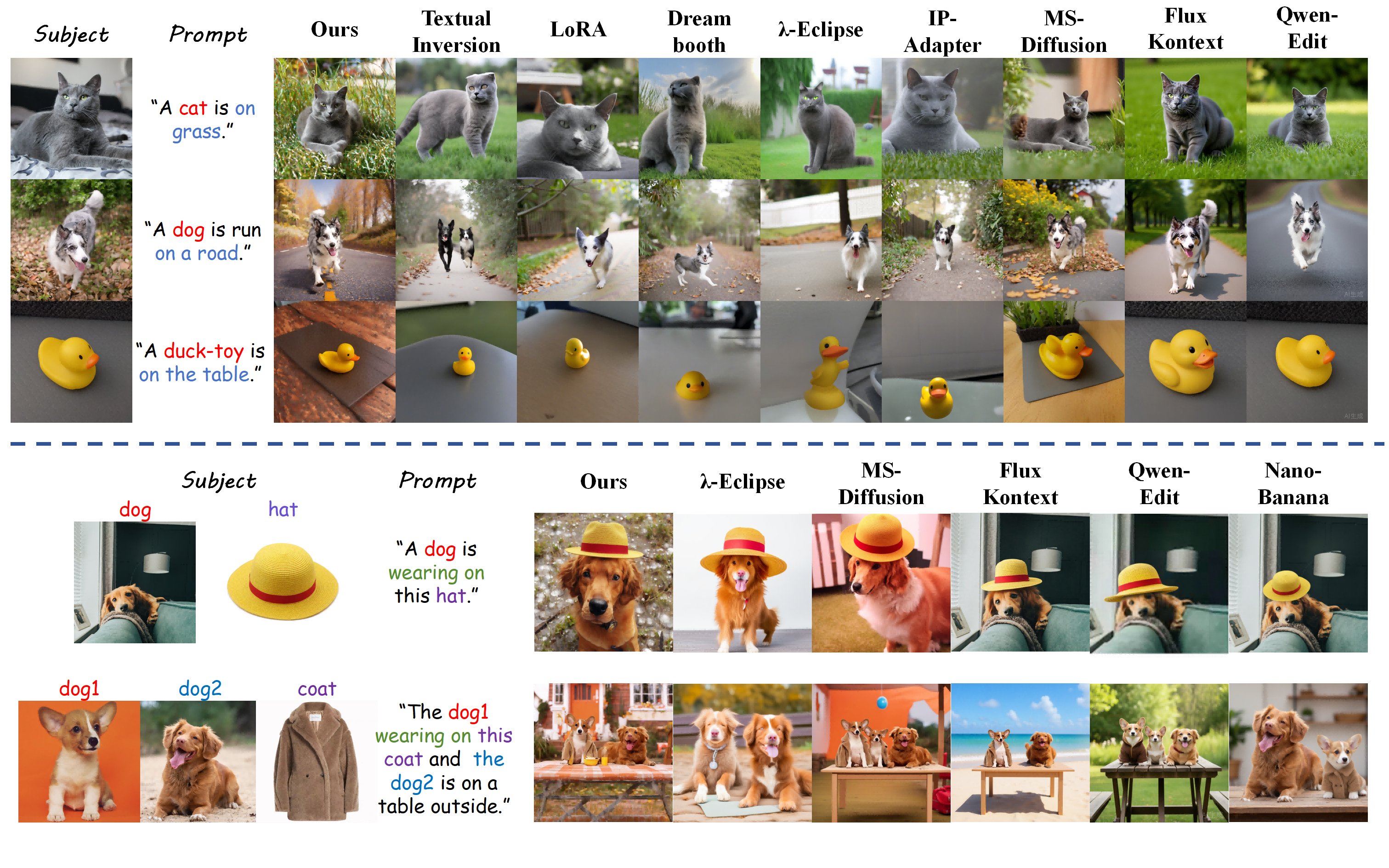}
  \caption{Additional single-frame image consistency visual experiments.}
  \label{fig:additional exp1}
\end{figure*}

\section{Additional Experiments}
\subsection{Single-frame Image Consistency(Addition)}
To further assess StoryTailor’s ability to preserve subject identity and obey action text in isolated frames, we ran additional single-frame consistency experiments on MSBench \cite{wang2024ms}. MSBench spans diverse multi-subject scenes. Combinations range from one to three subjects, with 100 long-form prompts per combination; each prompt is paired with per-subject reference images and grounding boxes. All images were generated at 1024×1024 using 50 sampling steps on a single RTX 4090 GPU. We compare StoryTailor against eight strong baselines that cover fine-tuning, adapter-based, in-context, and edit paradigms: LoRA\cite{hu2021lora}, DreamBooth\cite{ruiz2023dreambooth}, IP-Adapter\cite{ye2023ip}, MS-Diffusion\cite{wang2024ms}, FluxKontext\cite{labs2025flux}, $\lambda$-Eclipse\cite{patel2024lambda}, Qwen-Edit\cite{wu2025qwen}, and NanoBanana\cite{comanici2025gemini}. Metrics include CLIP-I for identity preservation, CLIP-T for text alignment, and Dino for subject similarity.

\textbf{Quantitative.} Quantitative results appear in Table \ref{tab:single_multi_partitioned_efficiency_global}. StoryTailor achieves a CLIP-I score of 0.849 in single subject cases. This outperforms MS-Diffusion at 0.824. The improvement stems from reduced background carryover through GCA. In multi-subject setups our method boosts CLIP-T by up to 15 percent. It reaches 0.414 compared to 0.340 for MS-Diffusion. AB-SVR amplification of action related embeddings drives this gain. It enhances interaction expressiveness without identity confusion. DINO-v2 and M-DINO remains strong. This indicates superior identity preservation.

\begin{table*}[t]
  \centering
  \caption{Single-frame Image Consistency on MSBench (micro-averaged means).
  Single-subject reports CLIP-I, DINO-v2, CLIP-T; Multi-subject reports CLIP-I, M-DINO, CLIP-T.
  \textbf{Best values are bold} and \underline{second-best values are underlined} for \emph{each column}.
  Inference Time and Peak GPU VRAM are overall efficiency metrics.}
  \label{tab:single_multi_partitioned_efficiency_global}
  \setlength{\tabcolsep}{4.5pt}
  \renewcommand{\arraystretch}{1.04}
  \begin{tabular*}{\textwidth}{@{\extracolsep{\fill}} l c c c c c c c c}
    \toprule
    & \multicolumn{3}{c}{\textbf{Single-subject}}
    & \multicolumn{3}{c}{\textbf{Multi-subject}}
    & \multicolumn{2}{c}{\textbf{Efficiency}} \\
    \cmidrule(lr{.5em}){2-4}\cmidrule(lr{.5em}){5-7}\cmidrule(lr{.5em}){8-9}
    \textbf{Method}
    & \textbf{CLIP-I} $\uparrow$
    & \textbf{DINO\mbox{-}v2} $\uparrow$
    & \textbf{CLIP-T} $\uparrow$
    & \textbf{CLIP-I} $\uparrow$
    & \textbf{M\mbox{-}DINO} $\uparrow$
    & \textbf{CLIP-T} $\uparrow$
    & \textbf{Time (s)} $\downarrow$
    & \textbf{VRAM (GB)} $\downarrow$ \\
    \midrule
    \multicolumn{9}{l}{\textit{Fine-tuning}} \\
    Textual Inversion    & 0.772 & 0.574 & 0.264 & --    & --    & --    & 3.51 & 13.73 \\
    LoRA                 & 0.795 & 0.591 & 0.270 & --    & --    & --    & 3.85 & 15.82 \\
    DreamBooth           & 0.802 & 0.589 & 0.309 & --    & --    & --    & 4.25 & 13.12 \\
    \addlinespace[1pt]
    \multicolumn{9}{l}{\textit{Adapters \& MLLM}} \\
    IP-Adapter           & 0.809 & 0.616 & 0.283 & --    & --    & --    & 4.82 & 19.39 \\
    MS-Diffusion         & 0.824 & 0.848 & 0.327 & 0.692 & 0.108 & 0.340 & 5.36 & 19.47 \\
    SSR-Encoder          & 0.813 & 0.799 & 0.331 & 0.723 & 0.104 & 0.311 & 7.49 & 18.82 \\
    $\lambda$-ECLIPSE    & 0.792 & 0.806 & 0.308 & 0.719 & 0.097 & 0.304 & 8.31 & 14.21 \\
    \addlinespace[1pt]
    \multicolumn{9}{l}{\textit{In-context}} \\
    FluxKontext          & \best{0.872} & 0.859 & 0.342 & \second{0.732} & 0.107 & 0.372 & 32.31 & 22.99 \\
    Qwen-Edit            & 0.841 & \best{0.902} & \second{0.373} & 0.714 & 0.108 & \second{0.396} & -- & -- \\
    NanoBanana           & \second{0.857} & \second{0.898} & 0.362 & \best{0.749} & \best{0.114} & 0.389 & -- & -- \\
    \midrule
    \textbf{StoryTailor (Ours)}
                         & 0.849 & 0.811 & \best{0.431} & 0.642 & \second{0.112} & \best{0.414} & 7.02 & 20.01 \\
    \bottomrule
  \end{tabular*}
\end{table*}

In summary, StoryTailor achieves the best CLIP-T for both single- and multi-subject because AB-SVR amplifies verb/interaction directions in text embeddings, GCA softly unbinds subject neighborhoods with a two-stage Gaussian to reduce overlap entanglement, and SFC keeps only transferable background cues while suppressing stale history—trading a bit of identity saturation for stronger text/action faithfulness. FluxKontext emphasizes global context and latent reuse, yielding strong identity (CLIP-I) but weaker verb grounding and much higher cost. Qwen-Edit (API editing) is strong on single-subject DINO-v2/CLIP-T, yet lacks structural cross-frame/ cross-subject constraints in crowded scenes, leading to identity–action trade-offs. Nano-Banana attains top multi-subject CLIP-I/M-DINO via tighter reference binding but remains conservative on verbs, hence lower CLIP-T. MS-Diffusion / SSR-Encoder / $\lambda$-ECLIPSE stabilize appearance similarity without explicit action boosting or overlap disentangling, so CLIP-T lags. IP-Adapter is weakest on verbs without added mechanisms. Fine-tuning methods (Textual Inversion/LoRA/DreamBooth) overfit limited views—identity is decent but actions under-expressed and they lack our/FluxKontext-style cross-frame memory and practical efficiency. Overall, the complementarity of GCA + AB-SVR + SFC prioritizes action faithfulness without freezing dynamics, while keeping identity and cost acceptable.

\textbf{Qualitative.} We characterize single-frame behavior as the number of subjects increases from one to three on MSBench, keeping resolution (1024×1024), steps (30), seeds, boxes, and per-subject references identical across methods; SFC is disabled to isolate single-frame effects. Baselines span LoRA, DreamBooth, IP-Adapter, MS-Diffusion, FluxKontext, $\lambda$-Eclipse, Qwen-Edit, and NanoBanana. Results is shown in Fig.\ref{fig:additional exp1}.

\subsection{Animals and Inanimate Objects Visual Narrative Task}
\begin{figure*}[t] 
  \centering
  \includegraphics[width=\linewidth]{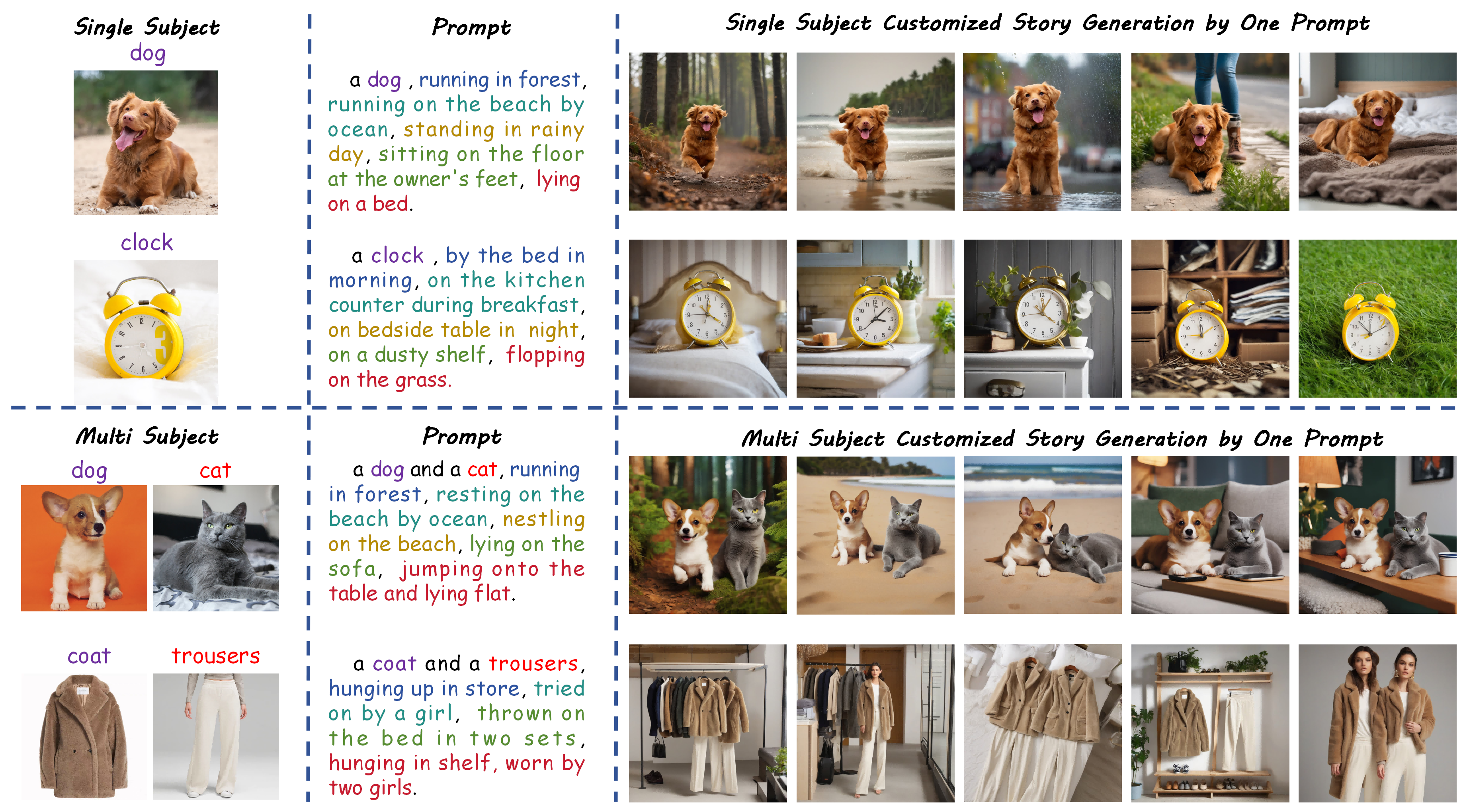}
  \caption{Additional animals and inanimate objects visual narrative task experiments.}
  \label{fig:additional exp6}
\end{figure*}
We illustrate the narrative ability of StoryTailor on both animals and inanimate objects, as shown in Fig. \ref{fig:additional exp6}. For each subject we provide one reference image and a single long-form prompt that lists a sequence of actions or scene descriptions. The pipeline decomposes the prompt into frame-level clauses and generates a visual story in which every clause corresponds to one image, while the subject identity and style remain stable.

For single-subject animals, the dog sequence covers actions such as “running in forest”, “running on the beach by ocean”, “standing in rainy day”, “sitting on the floor at the owner’s feet”, and “lying on a bed”. StoryTailor preserves the same dog across all frames, while backgrounds, lighting, and camera viewpoints vary in line with the described situations. The transition from forest to beach and finally to indoor scenes remains smooth. No extra dogs appear and the facial markings stay coherent, which shows that Gaussian-Centered Attention keeps the core identity stable while AB-SVR strengthens the verbs that drive the story.
For single-subject still life, we use a yellow clock and describe it “by the bed in morning”, “on the kitchen counter during breakfast”, “on bedside table in night”, “on a dusty shelf”, and “flopping on the grass”. The generated sequence depicts the same clock with consistent material and color under different contexts. The placements of the clock, supporting objects, and environment match the narrative, for example soft bedding, kitchen utensils, cardboard boxes, and outdoor grass. This suggests that our method can also build plausible stories for rigid objects that do not move on their own, by letting the environment and layout carry the narrative.

For multi-subject animals, the dog–cat sequence asks the pair to “run in forest”, “rest on the beach by ocean”, “nestle on the beach”, “lie on the sofa”, and “jump onto the table and lie flat”. StoryTailor produces a consistent pair whose fur color and body shape remain stable, while their relative poses and distances change in a natural way. The dog and cat share attention rather than collapsing into one hybrid creature, and occlusions such as the cat leaning on the dog are handled cleanly.
For multi-subject still life, the coat–trousers sequence describes “hanging up in store”, “tried on by a girl”, “thrown on the bed in two sets”, “hanging in shelf”, and “worn by two girls”. The results show one coherent outfit across different frames, from display racks to fitting scenes and lifestyle shots. The fabric tone and cut stay aligned with the reference images, while the composition adapts to the described usage scene, for example flat lay on the bed and two-person portrait at the end.

Overall, these examples indicate that StoryTailor can construct readable visual narratives for both animals and static objects under a single prompt. It maintains subject identity, enriches frame-wise actions and layouts, and arranges backgrounds so that the sequence reads as a short story rather than a set of unrelated images.

\begin{figure*}[t] 
  \centering
  \includegraphics[width=\linewidth]{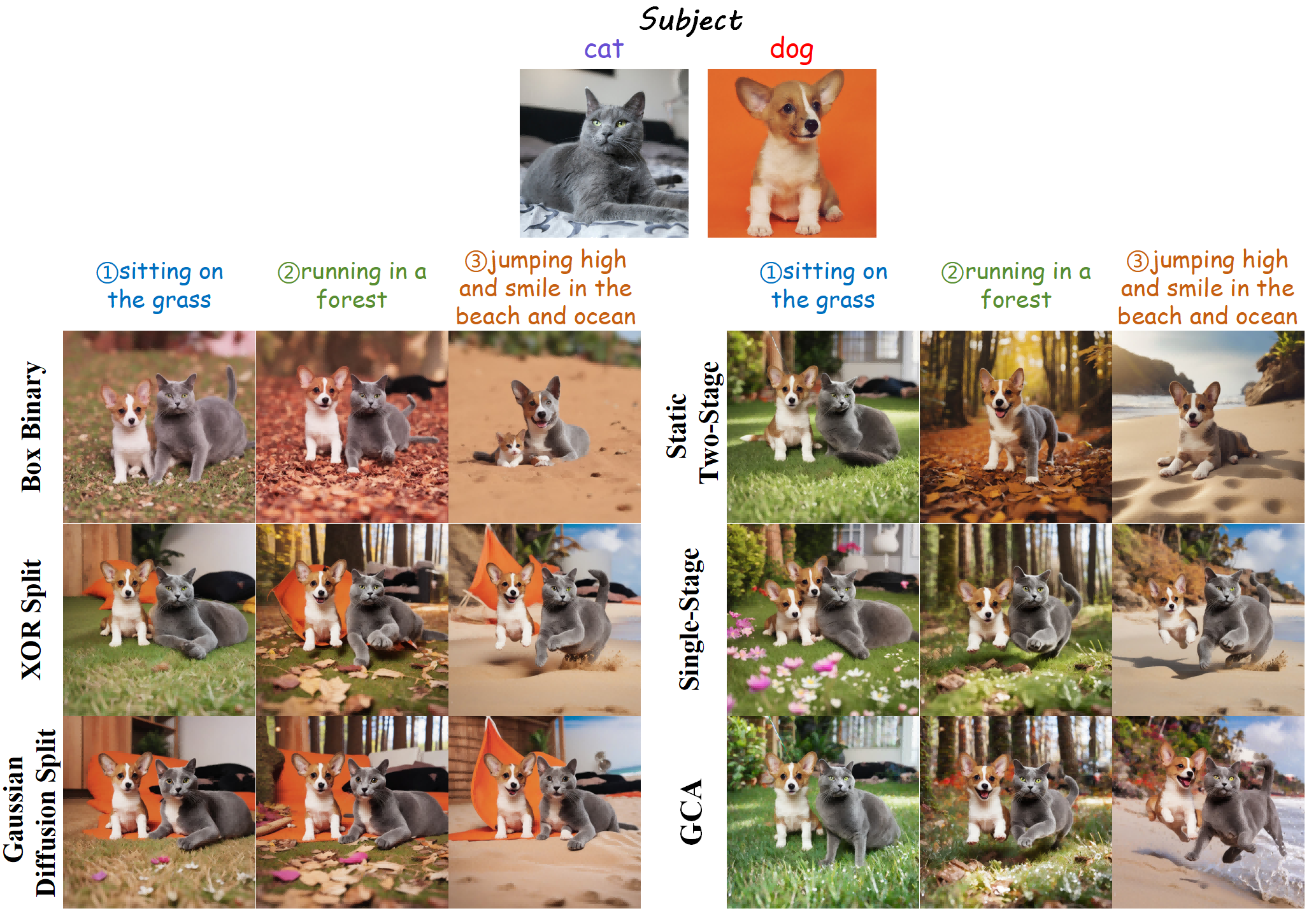}
  \caption{Additional GCA anisotropic comparison experiments.}
  \label{fig:additional exp2}
\end{figure*}

\begin{table*}[t]
  \centering
  \caption{GCA anisotropy ablation on multi-subject action scenes, 1024$\times$1024, 30 steps, shared seed, single 24\,GB GPU. 
  Metrics: CLIP-I$\uparrow$, CLIP-T$\uparrow$, background-drag$\downarrow$, subject confusion(Y/N), latency (s)$\downarrow$, peak VRAM (GB)$\downarrow$. 
  \emph{Best}=\best{bold}, \emph{second}=\second{underline}.}
  
  \setlength{\tabcolsep}{2.8pt}
  \resizebox{\linewidth}{!}{%
  \begin{tabular}{lcccccc}
    \toprule
    \textbf{Method} & \textbf{CLIP-I$\uparrow$} & \textbf{CLIP-T$\uparrow$} & \textbf{Bkg Drag$\downarrow$} & \textbf{Confusion(Y/N)} & \textbf{Latency$\downarrow$} & \textbf{Peak VRAM$\downarrow$} \\
    \midrule
    Box Binary                &0.815 & 0.285 & 13\% & Y & \best{6.23} &  \best{20.24}  \\
    XOR Split                 &0.837 & 0.279 & 22\% & N & \second{6.27} &  21.51  \\
    Gaussian Diffusion Split  &0.819 & 0.313 & 28\% & Y & 31.32&   22.01 \\
    Static Two-Stage          &\second{0.878} & \second{0.354} & 11\% & N & 7.13 &   20.58 \\
    Single-Stage              &0.856 & 0.349 & \second{6\%} & Y & 6.96 &   20.74 \\
    \textbf{GCA}              &\best{0.893} & \best{0.416} & \best{2\%} & N & 7.57 &  \second{20.44}  \\
    \bottomrule
  \end{tabular}%
  }
  \label{tab:gca_anisotropy_ablation_six}
\end{table*}

\begin{figure*}[t] 
  \centering
  \includegraphics[width=\linewidth]{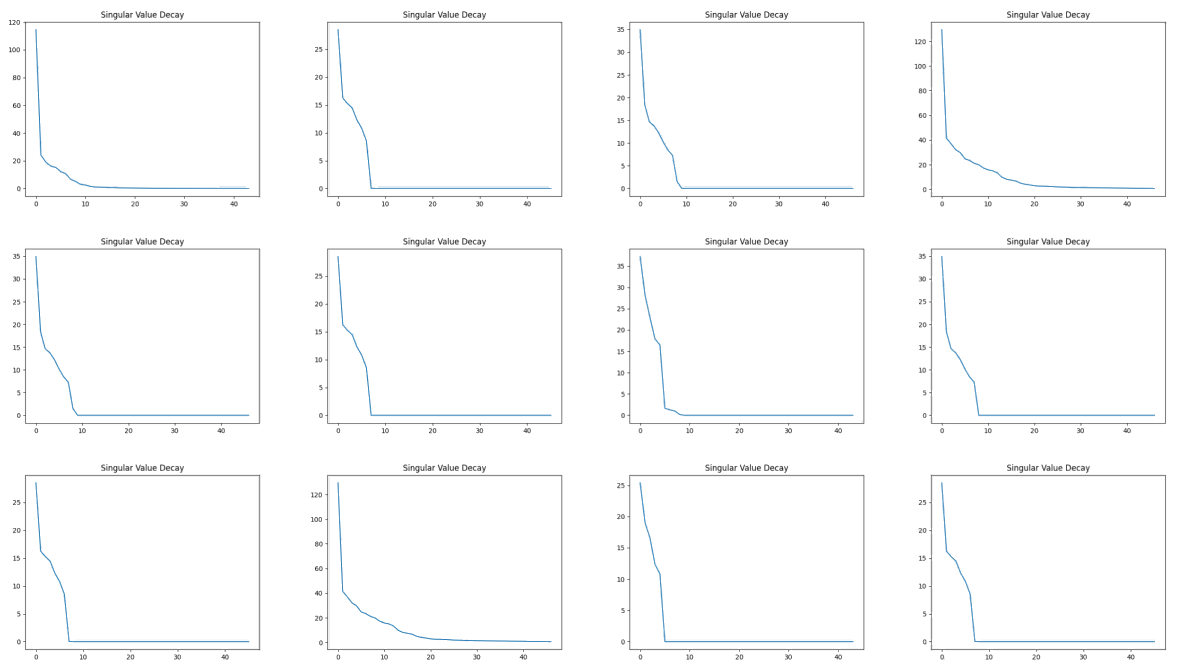}
  \caption{Some visual results of singluar value decay.}
  \label{fig:singular}
\end{figure*}

\section{GCA Anisotropic Experiments}
\subsection{Settings}

We perform an anisotropy ablation of GCA on multi-subject action scenes using a unified setup of 1024×1024 resolution, 30 sampling steps, shared seeds, and a single 24 GB GPU. We report micro-averaged CLIP-I and CLIP-T, measure background-drag ratio, log inference latency and peak VRAM, and flag any subject confusion. GCA is kept as the core while the other module is kept the same. To avoid unverified subject confusion and entanglement, we selected non-interactive actions and set the grounding boxes to [[[0.2, 0.2, 0.6, 0.6], [0.45, 0.20, 0.8, 0.60]]], resulting in an overlap region occupying approximately 12\% of the total image area, to validate the methods' effectiveness in subject decoupling. We compare the following six strategies:

\textbf{Box Binary}: in-box binary mask with background dummy as a geometric baseline.

\textbf{XOR Split}: pure XOR hard disentangling for overlaps with direct separation in conflict regions.

\textbf{Gaussian Diffusion Split}: pseudo masks from diffusion-time aggregated text attention with Gaussian-strength adaptive splitting over overlaps and map-derived bias injected to image-side logits.

\textbf{Static Two-Stage}: two-stage Gaussian via long-side split and dual centroids with fixed radii that do not vary with semantics.

\textbf{Single-Stage}: single-stage isotropic Gaussian inside the box with soft sharing in overlaps dynamically driven by attention heatmaps.

\textbf{GCA}: two-stage anisotropic Gaussian via long-side split and dual centroids with radii dynamically driven by attention heatmaps.

\subsection{Analysis}
The Fig.\ref{fig:additional exp2} contrasts baseline mask methods left: Binary Box, XOR Split, Gaussian Diffusion Split with two-stage variants right: Static Two-Stage, Single-Stage, GCA in generating multi-subject dog-cat scenes across three actions. Baselines exhibit identity swaps, static poses, and background inconsistencies e.g., leaked elements in jumping scenes, while GCA produces vivid interactions, accurate actions e.g., dynamic runs and smiles, and stable evolving backgrounds, aligning with the paper's claims of superior overlap resolution and action richness via anisotropic Gaussians. Table. \ref{tab:gca_anisotropy_ablation_six} experimental results reveal that binary box masking causes severe identity confusion and background drag in multi-subject scenes, leading to stiff action expressions and lackluster interactions, while Gaussian diffusion splitting offers slight relief from drag but escalates computational demands. Static two-stage variants start to mitigate identity leakage and improve action coherence, yet they are hampered by rigid boundaries that provoke local conflicts. Single-stage dynamic Gaussian outperforms prior strategies, but action attributes remain insufficiently prominent; two-stage dynamic Gaussian GCA excels markedly, producing vivid interactions, precise actions, and gradually stable backgrounds for superior overall narrative fluency. The rationale stems from GCA's anisotropic Gaussians dynamically centering on subject cores to soften overlap boundaries and avert confusion, paired with AB-SVR's reinforcement of action embedding directions to infuse behavioral diversity, and SFC's selective retention of background semantics over frozen history to secure cross-frame continuity without curtailing subject dynamism. This modular synergy deftly resolves the zero-shot pipeline's inherent tensions.

\begin{figure*}[t]
  \centering
  \includegraphics[width=\linewidth]{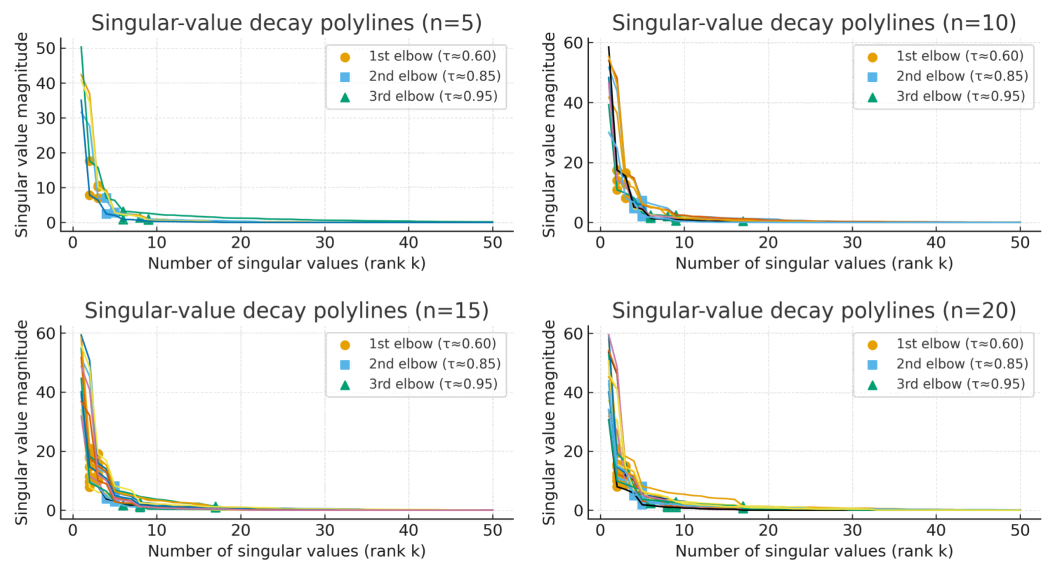}
  \caption{
  Singular-value decay polylines for different numbers of sampled prompts. Each subplot visualizes $n\!=\!5,10,15,20$ text embeddings, respectively. For every curve we plot the singular-value spectrum and highlight the three largest energy drops with markers: the first elbow (circles, $\tau\!\approx\!0.60$), the second elbow (squares, $\tau\!\approx\!0.85$), and the third elbow (triangles, $\tau\!\approx\!0.95$). The knees cluster into three bands, corresponding to subject-, action-, and background-related energy regions in the text-embedding space.
  }
  \label{fig:svd_decay_grid}
\end{figure*}

\begin{figure*}[t] 
  \centering
  \includegraphics[width=\linewidth]{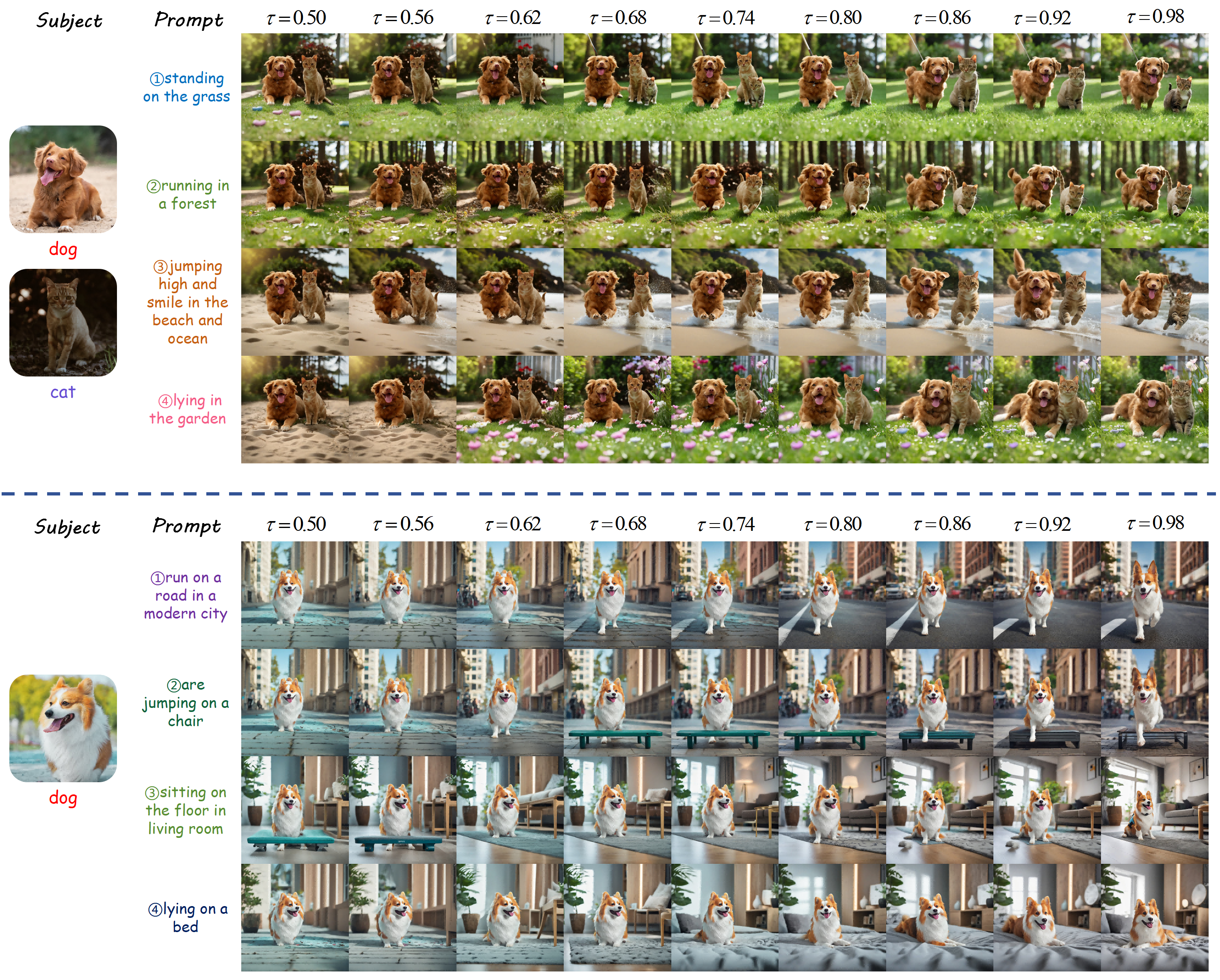}
  \caption{Singluar value energy $\tau$ setting experiments on single- and multi-subject tasks.}
  \label{fig:additional exp3}
\end{figure*}

\begin{table}[t]
  \centering
  \caption{Statistical distribution of knee positions $k$ across cumulative energy thresholds $\tau$ in 200 samples.}
  \label{tab:tau_knee_distribution}
  \setlength{\tabcolsep}{7pt}     
  \renewcommand{\arraystretch}{1.1} 
  \small                            

  \begin{tabular*}{\columnwidth}{@{\extracolsep{\fill}}lcccc}
    \toprule
    $\tau$ range & $k_{\min}$ & $k_{\max}$ & $k_{\text{avg}}$ & Knees \\
    \midrule
    0.50--0.60 &  2 &  3 &  2.1 &  18 \\
    0.61--0.70 &  2 &  4 &  2.8 &  32 \\
    0.71--0.80 &  3 &  6 &  4.1 &  41 \\
    0.81--0.90 &  3 &  8 &  5.0 &  53 \\
    0.91--0.95 &  5 & 18 & 10.2 &  37 \\
    0.96--0.99 &  8 & 39 & 19.5 &  19 \\
    \bottomrule
  \end{tabular*}
\end{table}

\section{Singluar Value Energy Experiment}
To understand how the singular-value energy of each sentence in SVR is distributed over the semantic subspace of the text embedding, we first feed multiple prompts into the text encoder to obtain their embeddings. For each individual sequence of semantic vectors, we perform thin SVD and visualize the decay curves of singular-value energy, which allows us to inspect how different energy bands contribute to the text-embedding subspace. Fig. \ref{fig:singular}  presents some visual results of singular value decay curves together with statistics of the detected energy knees.

In multi-frame visual narrative generation, we first split the global story prompt into frame-level paragraphs and perform thin SVD on the text embedding of each frame. The frame-wise embedding matrix is factorized along the “feature–token” axes as
\begin{equation}
    X = U \Sigma V^T
\end{equation}
where $\Sigma$ is a diagonal matrix and each singular value $\sigma_i$ represents the scaling factor or energy contribution of the $i$-th principal direction. The squared singular value $\sigma_i^{2}$ corresponds to the variance captured by that component, i.e., the amount of data variability along this direction. Large singular values therefore encode the dominant transformations that carry most of the information in the embedding, whereas small singular values are associated with minor components or noise. At the text-encoding stage, this spectrum induces a natural semantic hierarchy: leading singular values are dominated by subject entities and core attributes, mid-range singular values are more aligned with action details and interaction relations, and tail singular values mostly reflect background style and fine-grained modifiers. A key use of SVD in our setting is low-rank approximation: by truncating smaller singular values, we reconstruct $\mathbf{X}$ with far fewer dimensions while losing only limited information.

We focus on detecting elbows in the singular-value decay curves, because these points mark the boundary between dominant and secondary information. Many real-world matrices, including image features and text embeddings, exhibit a low-rank nature, where the effective rank is much smaller than the nominal dimensionality. In the multi-frame setting, as the number of frames increases, frame-level paragraphs interfere with each other in the embedding space: semantics of non-target frames are suppressed but not completely removed, so residual signals from other frames remain in the principal spectrum. To obtain a cleaner representation for the current frame, we therefore rely on the structure of the singular-value decay. Empirically, the first pronounced energy drop corresponds to backbone subject information, the second major drop aggregates action and interaction semantics, and the third drop mainly captures background and scene style, while the remaining long tail is largely cross-frame residue and noise. Based on this interpretation, we apply low-rank approximation to select singular values within the desired energy bands, retaining subject- and action-related components for the current frame while discarding as much inter-frame noise and redundancy as possible.

As illustrated in Fig.~\ref{fig:svd_decay_grid}, we further visualize the singular-value decay patterns of frame-level text embeddings. For each setting we randomly sample $n\!\in\!\{5,10,15,20\}$ frame-level paragraphs, compute their embeddings, and plot the sorted singular values $\sigma_i$ against the rank $k$. On every polyline, we mark the three strongest energy drops as the first, second, and third elbows, which approximately correspond to cumulative energy thresholds $\tau\!\approx\!0.60$, $\tau\!\approx\!0.85$, and $\tau\!\approx\!0.95$, respectively. As $n$ increases, these knees consistently cluster into three bands along the $k$ axis, confirming that the spectra of text embeddings exhibit a stable low-rank hierarchy: the first elbow captures subject-centric backbone semantics, the second elbow aggregates action and interaction information, and the third elbow is dominated by background and style. This empirical behavior supports our subsequent choice of $\tau$ bands
for AB-SVR, where we explicitly operate on these energy regions to retain subject and action components while suppressing residual background and cross-frame noise.

Figure~\ref{fig:additional exp3} further provides a qualitative view of how the cumulative-energy threshold $\tau$ in AB-SVR affects the generated semantics. We fix the subjects (dog or dog+cat) and frame-level prompts, and sweep $\tau$ from $0.50$ to $0.98$. When $\tau$ is too small (e.g., $0.50$ or $0.56$), only a few leading components are retained: subjects remain roughly identifiable, but actions become attenuated and often collapse into static poses, while backgrounds are under-specified and frequently drift across frames. As $\tau$ enters the mid-range band around $0.74$--$0.80$, the reconstructed text embeddings recover more action- and relation-related directions: dogs and cats not only appear with stable identities, but also execute the requested actions (running, jumping, lying) more faithfully, and the surrounding context becomes more coherent without overwhelming the subjects. Pushing $\tau$ further towards $0.92$ and $0.98$ introduces a large portion of background- and style-dominated components, which strengthens texture and scenery but also re-couples subjects and background; in multi-
subject scenes this occasionally leads to pose over-regularization or subtle identity blending. These trends are consistent with the spectral analysis in Fig.~\ref{fig:svd_decay_grid}: the first elbow mainly supports subject cores, the second elbow enriches action semantics, and the third elbow starts to absorb background energy, making $\tau\approx0.80$ a reasonable operating point that balances identity fidelity, action expressiveness, and background consistency.

\section{Long-frame Narrative Effects}

\begin{figure}[t] %
  \centering
  \includegraphics[width=\columnwidth]{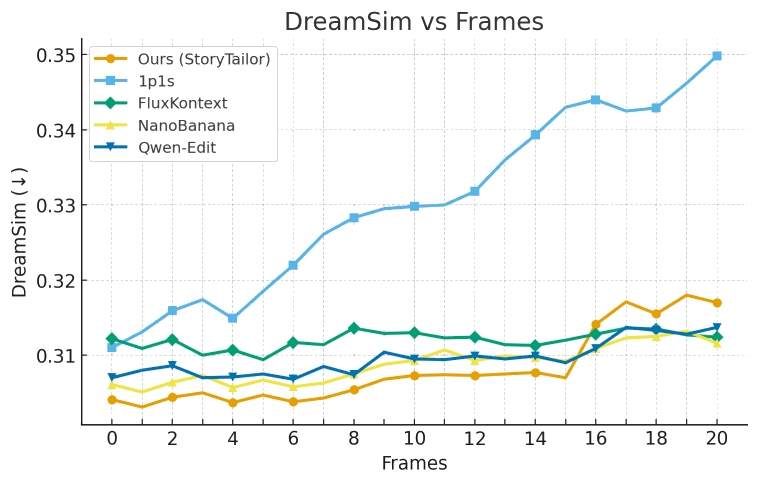} 
  \caption{DreamSim vs.\ sequence length on MSBench (lower is better).}
  \label{fig:sup_dreamsim}
\end{figure}

\begin{figure}[t] %
  \centering
  \includegraphics[width=\columnwidth]{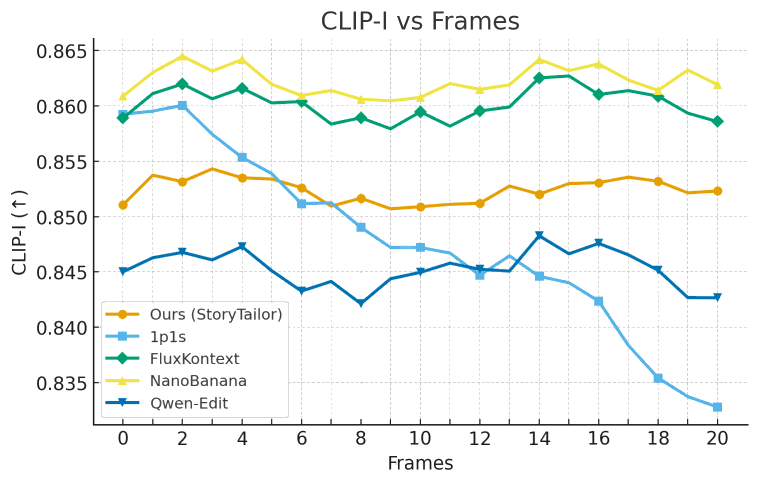} 
  \caption{CLIP-I vs.\ sequence length on MSBench (higher is better).}
  \label{fig:sup_clipi}
\end{figure}

\begin{figure}[t] %
  \centering
  \includegraphics[width=\columnwidth]{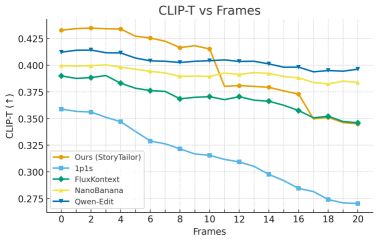} 
  \caption{CLIP-T vs.\ sequence length on MSBench (higher is better).}
  \label{fig:sup_clipt}
\end{figure}

We further study how StoryTailor behaves on long-horizon narratives using the MSBench prompts. For each method, we generate up to 20 frames per story under identical prompts, reference images, bounding boxes, seeds, and sampling steps, and compute DreamSim (↓), CLIP-I (↑), and CLIP-T (↑) for every frame. The curves in Fig.~\ref{fig:sup_dreamsim}--\ref{fig:sup_clipt} are averaged over all sequences.

In terms of DreamSim (Fig.~\ref{fig:sup_dreamsim}), StoryTailor maintains the lowest perceptual distance over almost the entire horizon. Our curve starts around 0.25 and only increases by about 0.03 when moving to 20-frame stories, whereas 1p1s rises by roughly 0.16 over the same range and thus accumulates substantial temporal noise. FluxKontext, NanoBanana, and Qwen-Edit stay in the middle: their DreamSim values are more stable than 1p1s but remain consistently higher than StoryTailor at long horizons. This suggests that the combination of Gaussian-Centered Attention, Action-Boost SVR, and Selective Forgetting Cache effectively dampens error accumulation and avoids the “drift into chaos’’ often observed in long sequences.

CLIP-I results (Fig.~\ref{fig:sup_clipi}) show that StoryTailor keeps identity fidelity in a narrow and visually acceptable band around 0.85, with a total variation below 0.01 across all frame indices. In contrast, 1p1s exhibits a
clear downward trend and loses about 0.03 in CLIP-I as sequences become longer, indicating gradual identity erosion. In-context baselines (FluxKontext, NanoBanana, Qwen-Edit) retain slightly higher CLIP-I than ours, but qualitative inspection reveals that they do so by producing more static,pose-conservative content. Our method therefore trades a small amount of identity tightness—still acceptable to human observers—for richer and more persistent motion.

For CLIP-T (Fig.~\ref{fig:sup_clipt}), StoryTailor consistently achieves the best text–image alignment. Our CLIP-T starts above 0.43 and remains above 0.40 throughout the 20-frame horizon. By comparison, 1p1s drops by almost 0.09,
showing strong long-frame degradation; FluxKontext also decreases notably with length, while NanoBanana and Qwen-Edit decline more gently but stay below StoryTailor for most frames. The relatively flat CLIP-T curve of StoryTailor indicates that Action-Boost SVR repeatedly recenters verb and interaction directions at each frame, and that Selective Forgetting Cache limits the impact of stale history on new textual intents. Overall, these long-frame results confirm that StoryTailor can sustain narrative dynamics over extended horizons: perceptual drift is controlled, identity remains stable, and action semantics stay well aligned with the story text.

\begin{figure*}[t] 
  \centering
  \includegraphics[width=\linewidth]{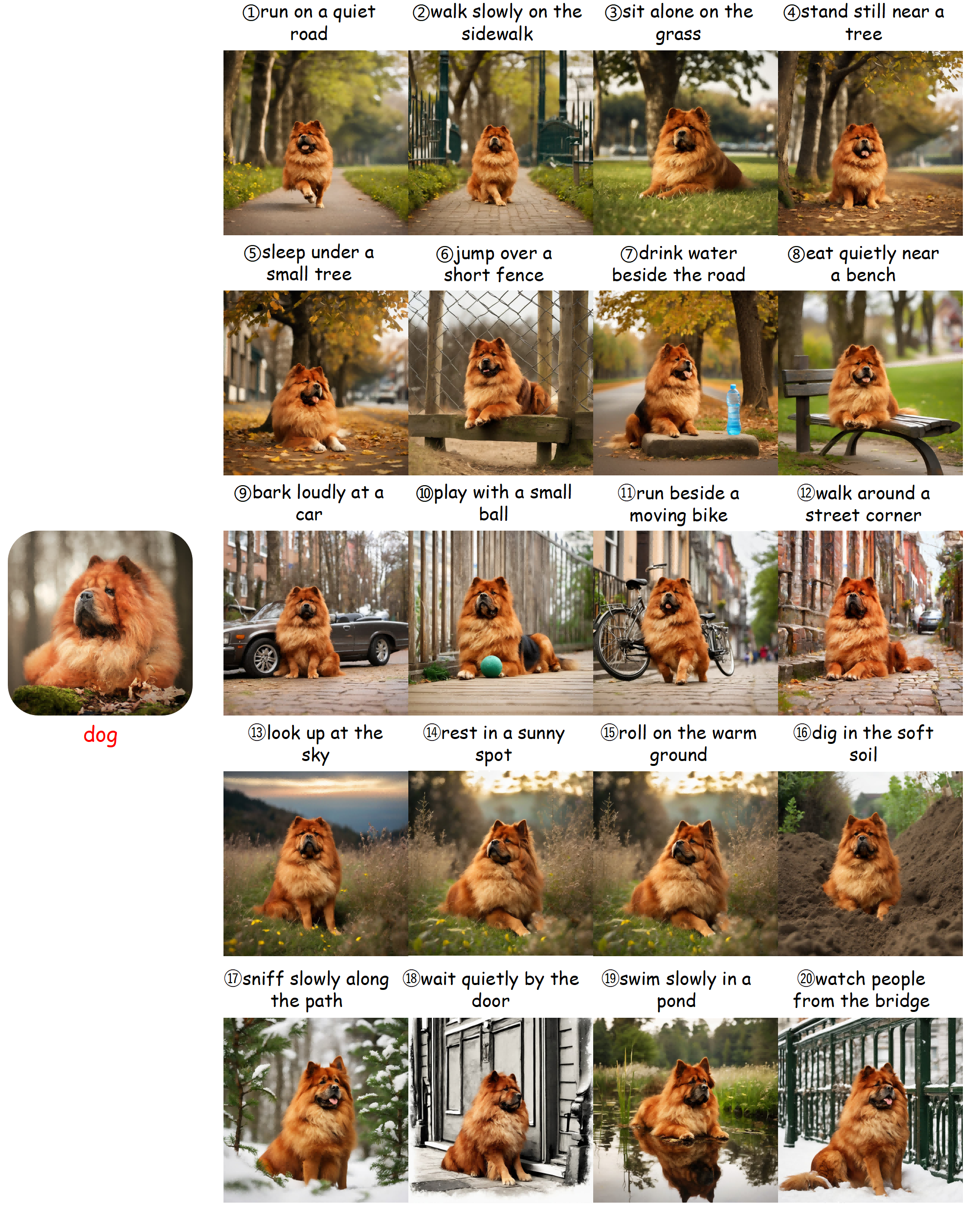}
  \caption{Long-frame narrative effect experiment on single-subject tasks (a 20-frame example).}
  \label{fig:additional exp4}
\end{figure*}

\begin{figure*}[t] 
  \centering
  \includegraphics[width=0.98\linewidth]{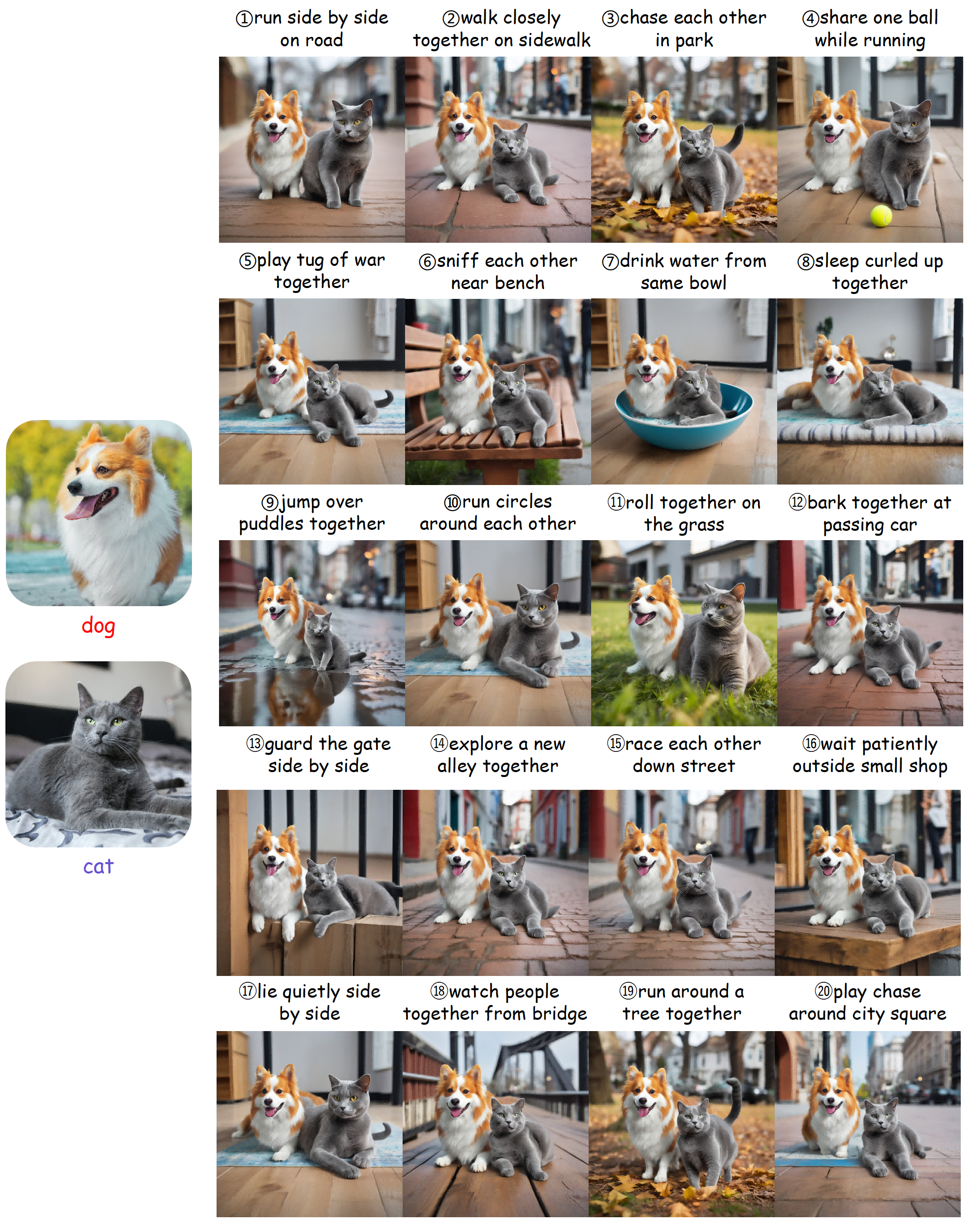}
  \caption{Long-frame narrative effect experiment on multi-subject tasks (a 20-frame example).}
  \label{fig:additional exp5}
\end{figure*}

Fig. \ref{fig:additional exp4} shows a 20-frame single-subject narrative of a dog moving through streets, parks, and seasonal scenes. The dog identity stays highly consistent despite large changes in viewpoint, background, weather, and even style (e.g., snow and black-and-white frames). The sequence covers a diverse set of actions, such as walking, sitting, resting in a sunny spot, digging in the soil, swimming slowly in a pond, and waiting quietly by the door, forming a plausible progression of everyday activities. Nevertheless, long-frame effects are also visible here: in later frames some prompts with very specific verbs are rendered as more generic “standing” or “sitting” poses, and the distinction between neighboring actions becomes softer than in early frames. Compared with the multi-subject case, identity and layout are easier to preserve, but fine-grained verb expression still tends to be partially realized when the story extends to 20 frames, which again aligns with the slight long-horizon drop in CLIP-T.

Fig. \ref{fig:additional exp5} presents a 20-frame narrative with two subjects, a brown dog and a gray cat. The pair traverse the same urban environment while performing a series of explicitly prompted joint actions, including running side by side, sharing a ball, playing tug of war, drinking from the same bowl, and waiting together outside a small shop. Across all frames, both identities remain stable and well separated, the dog–cat layout stays plausible, and the background evolves smoothly along a consistent city route, indicating strong subject fidelity and cross-frame continuity. At the same time, a clear long-frame effect appears in the later part of the sequence: some fine-grained interaction verbs (e.g., “guard the gate,” “watch people from the bridge”) are only partially realized or simplified into visually similar side-by-side poses, and the difference between consecutive actions becomes less pronounced. This matches the mild CLIP-T decline with increasing frame index and shows that, under multi-subject interaction, expressing every detailed verb perfectly becomes more difficult as the narrative horizon grows longer.

\section{User Study}

\subsection{Protocol}
\begin{figure*}[t] 
  \centering
  \includegraphics[width=0.89\linewidth]{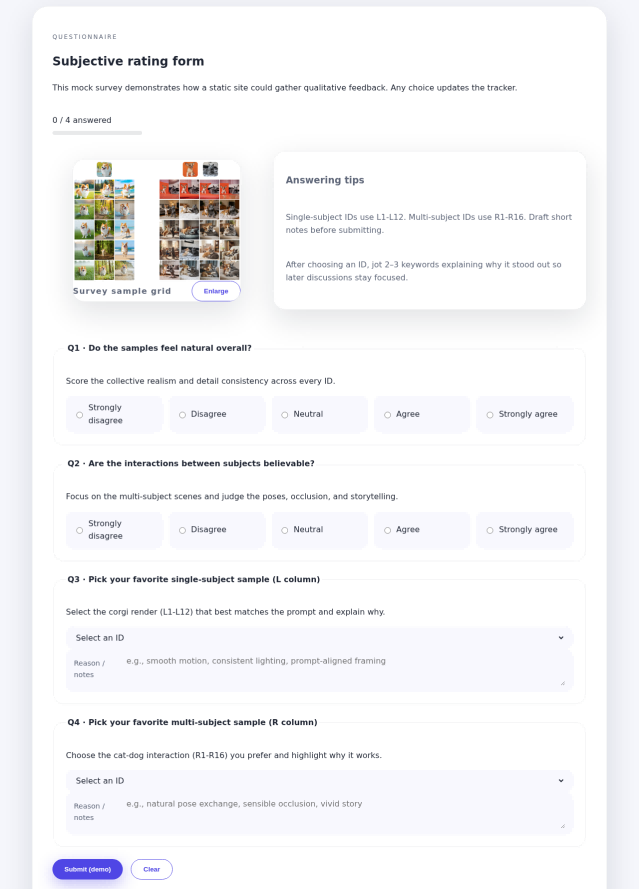}
  \caption{The web questionaire of StoryTailor about visual narrative tasks for users.}
  \label{fig:questionaire}
\end{figure*}

We deploy a web-based survey interface, shown in Fig. \ref{fig:questionaire}. Each trial displays a grid of images generated from one long prompt by several anonymous methods: StoryTailor and the baselines in Sec. 4.2–4.3. Single-subject trials use twelve dog-action scenes indexed L1–L12. Multi-subject trials use sixteen cat–dog interaction scenes indexed R1–R16, such as hugging, dancing, and hand-over. Each column corresponds to one model; column order and prompt order are shuffled for every participant, and model names never appear, giving a double-blind setting.

Participants answer four questions for each grid.

Q1 asks how natural and internally consistent the full set looks. Ratings follow a five-point Likert scale that ranges from “Strongly disagree” to “Strongly agree”.

Q2 asks whether multi-subject interactions look believable, with emphasis on poses, occlusion, and short-story coherence.

Q3 asks participants to pick a favourite single-subject image ID from L1–L12 and give a short reason, for instance smooth motion, stable lighting, or good alignment with the prompt.

Q4 asks for a favourite multi-subject image ID from R1–R16 and a short explanation, such as natural pose exchange, sensible occlusion, or vivid storytelling.

We recruit 100 volunteers, mainly graduate students and researchers in vision and graphics. Every participant completes all single- and multi-subject trials.

\subsection{Analysis and Findings.}
\begin{figure*}[t] %
  \centering
  \includegraphics[width=0.75\linewidth]{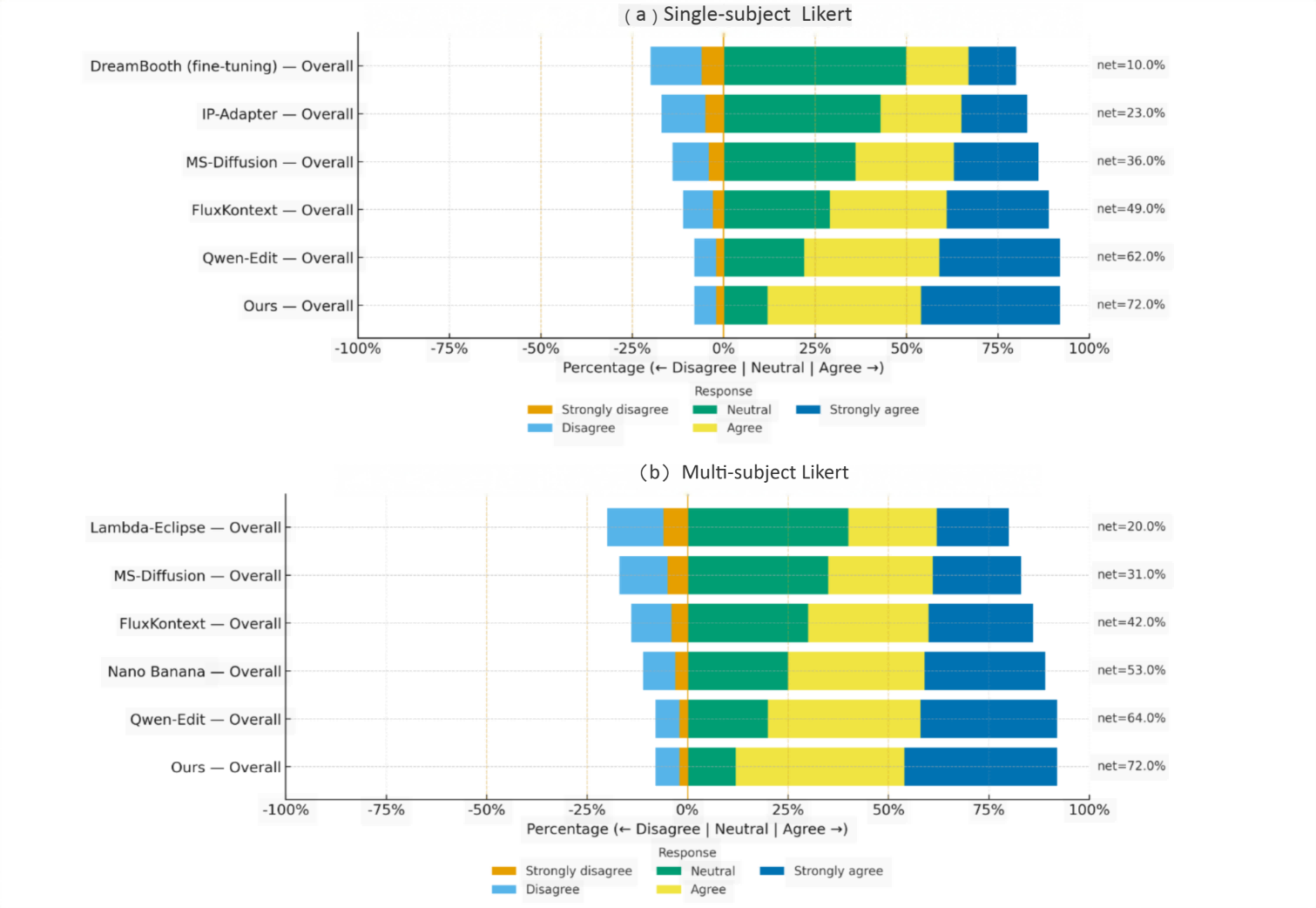} 
  \caption{The Likert scores for single-subject(a) and multi-subject(b) tasks for user study}
  \label{fig:likert_scores}
\end{figure*}

\textbf{Single-subject ratings.}
Fig. \ref{fig:likert_scores}(a) shows the Likert scores for single-subject prompts. Our method reaches a net agreement of about 72/\%, clearly ahead of all baselines. Qwen-Edit is the strongest competitor with roughly 62\% net agreement. FluxKontext follows at about 49\%, while MS-Diffusion, IP-Adapter, and DreamBooth obtain around 36\%, 23\%, and 10\%, respectively. The distribution of bars also shifts to the right for our model: most responses fall into “agree” or “strongly agree”, and the fraction of negative ratings is the smallest among all methods. These trends suggest that StoryTailor preserves identity well and expresses the action cues in the prompt more faithfully, even when only one subject is present.

\textbf{Multi-subject ratings.}
For multi-subject scenes in Fig. \ref{fig:likert_scores}(b), the gap widens. Our method again achieves about 72\% net agreement, while Qwen-Edit reaches 64\%. Nano Banana and FluxKontext obtain roughly 53\% and 42\%, and MS-Diffusion and Lambda-Eclipse lag behind with about 31\% and 20\%. Reviewers report that our results have cleaner backgrounds, fewer collisions between subjects, and more coherent pose progressions across the sequence. Baselines more often receive neutral or negative scores due to background drag, attribute spill, or awkward contact between subjects.

\begin{figure*}[t] %
  \centering
  \includegraphics[width=0.7\linewidth]{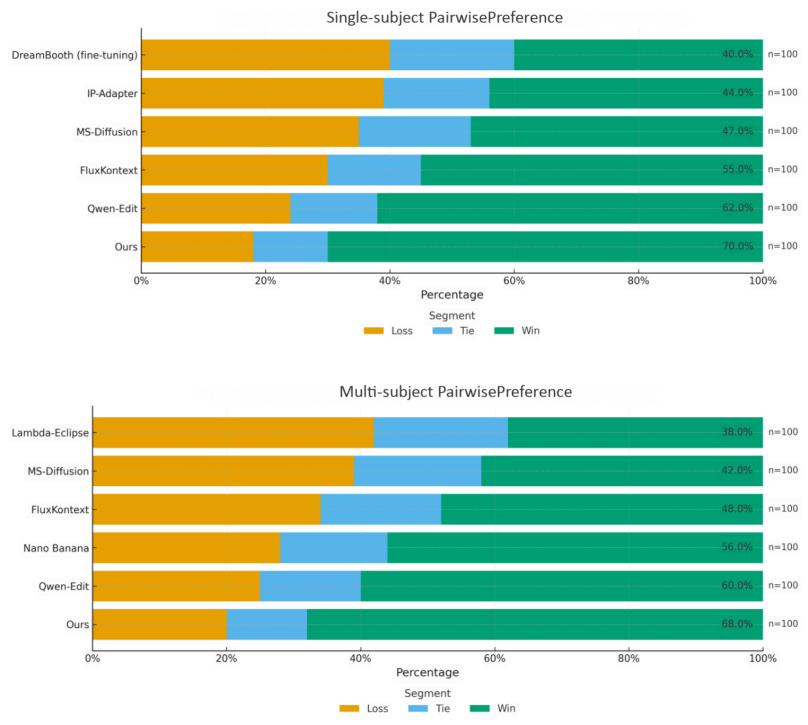} 
  \caption{The pairwise preference for single-subject(a) and multi-subject(b) tasks for user study}
  \label{fig:pairwise}
\end{figure*}

\textbf{Pairwise preference.}
Pairwise comparisons in Fig. \ref{fig:pairwise} confirm these findings. On single-subject prompts, our method attains an overall preference score of about 70\%, while Qwen-Edit reaches 62\%, FluxKontext 55\%, and MS-Diffusion, IP-Adapter, and DreamBooth stay between 40\% and 47\%. On multi-subject prompts, our method still leads with about 68\% preference, followed by Qwen-Edit with 60\%, Nano Banana with 56\%, FluxKontext with 48\%, and MS-Diffusion and Lambda-Eclipse with 42\% and 38\%. In qualitative comments, participants frequently describe our images as “more natural interactions” and “less confusing when subjects are close to each other”.


StoryTailor attains the highest mean opinion score for both overall naturalness and interaction believability, with the clearest gains in multi-subject scenes. Participants often describe our outputs as having cleaner backgrounds, more natural pose exchanges, and smoother frame-to-frame progression. Baselines more frequently show background drag, attribute leakage, or awkward contact between subjects. In pairwise preference, StoryTailor wins the majority of comparisons against MS-Diffusion, FluxKontext, Qwen-Edit, NanoBanana and other strong baselines, especially in crowded or heavily occluded scenes. These subjective trends match the quantitative metrics: GCA and SFC reduce background carry-over and close-range confusion, while AB-SVR sharpens verb and interaction cues, leading to visual narratives that human viewers consistently prefer.
\section{Limitations}
Although StoryTailor improves action expression and multi-subject coherence under a 24 GB budget, several limitations remain. First, our design and evaluation are tied to SDXL with MS-Diffusion-style multi-subject conditioning and CLIP-family metrics, so generalization to other backbones, styles, and prompt distributions is not fully verified. Second, we mainly study short to medium narratives with 2–20 frames; as the sequence length grows, text embedding noise accumulates and CLIP-T gradually degrades, indicating that long-range story structure and very long narratives are not yet fully handled. Third, the current GCA and AB-SVR hyperparameters are hand-tuned for a limited set of subjects and actions, and performance can drop under extreme poses, heavy occlusions, dense crowds, or very loose bounding boxes. Fourth, SFC assumes moderate background continuity and may either over-smooth or under-propagate context in scenes with abrupt layout changes, fast camera motion, or strong lighting transitions. Finally, we rely on a small-scale user study and automatic metrics that only approximate human judgment, so a more systematic perceptual evaluation and task-specific benchmarks are left for future work.
\section{Social Impacts}
By making multi-subject, action-rich visual narratives feasible on a single commodity GPU, StoryTailor can support positive applications in education, digital heritage, pre-visualization, assistive storytelling, and personal creativity, especially for users without access to large GPU clusters. At the same time, the system operates on personal reference images and identity-consistent generation, which raises clear risks around privacy, consent, and potential misuse for deepfakes, harassment, or misleading narrative content. Our pipeline does not alter or weaken the safety mechanisms of the underlying diffusion backbone, but it may inherit and amplify dataset biases, stereotypes, or unequal representation patterns present in the pretrained models. Generated narratives could unintentionally reinforce cultural or gender stereotypes in how actions, roles, and relationships are depicted, particularly in multi-subject scenes. We therefore recommend deploying StoryTailor together with permission management for reference images, content filters, and human review in sensitive domains, and encourage future work on fairness-aware training data, controllable safety constraints, and tools that help users detect and mitigate harmful or deceptive generations.

\end{document}